\documentclass[10pt,journal,compsoc]{IEEEtran}
\ifCLASSOPTIONcompsoc
  \usepackage[nocompress]{cite}
\else
  \usepackage{cite}
\fi

\usepackage[pdftex]{graphicx}
\usepackage{makecell}
\usepackage{booktabs}
\usepackage{multirow}
\usepackage{amsfonts}
\usepackage{amsmath}
\usepackage{mathrsfs}
\usepackage{bm}
\usepackage{cases}
\usepackage{balance}
\usepackage{algorithm}
\usepackage{algorithmic}

\newtheorem{myDef}{Definition}
\newtheorem{myExam}{Example}
\usepackage{enumerate}

\begin{document}
\title{What is Event Knowledge Graph: A Survey}
\author{Saiping~Guan,~Xueqi~Cheng,~Long~Bai,~Fujun~Zhang,~Zixuan~Li,~Yutao~Zeng,~Xiaolong~Jin, and~Jiafeng~Guo
\IEEEcompsocitemizethanks{\IEEEcompsocthanksitem 
Saiping Guan, Xueqi Cheng, Long Bai, Fujun Zhang, Zixuan Li, Xiaolong Jin, and Jiafeng Guo are with the School of Computer Science and Technology, University of Chinese Academy of Sciences, Beijing 100049, China, and also with the CAS Key Lab of Network Data Science and Technology, Institute of Computing Technology, Chinese Academy of Sciences, Beijing 100864, China. E-mail: \{guansaiping,cxq,bailong18b,zhangfujun19s,lizixuan,jinxiaolong,guojia- feng\}@ict.ac.cn.
}
\IEEEcompsocitemizethanks{\IEEEcompsocthanksitem 
Yutao~Zeng is with the Platform and Content Group, Tencent, Beijing, 100080, China. E-mail: tedzeng@tencent.com.}

\thanks{Manuscript received 26 Nov. 2021; revised 10 Apr. 2022; accepted 22 May 2022 (Corresponding authors: Xueqi Cheng, Xiaolong Jin, and Jiafeng Guo).}}

\markboth{Journal of \LaTeX\ Class Files,~Vol.~14, No.~8, August~2015}%
{Shell \MakeLowercase{\textit{Guan et al.}}: What is Event Knowledge Graph: A Survey}

\IEEEtitleabstractindextext{%
\begin{abstract}
Besides entity-centric knowledge, usually organized as Knowledge Graph (KG), events are also an essential kind of knowledge in the world, which trigger the spring up of event-centric knowledge representation form like Event KG (EKG). It plays an increasingly important role in many downstream applications, such as search, question-answering, recommendation, financial quantitative investments, and text generation. This paper provides a comprehensive survey of EKG from history, ontology, instance, and application views. Specifically, to characterize EKG thoroughly, we focus on its history, definition, schema induction, acquisition, related representative graphs/systems, and applications. The development processes and trends are studied therein. We further summarize prospective directions to facilitate future research on EKG. 
\end{abstract}

}

\maketitle
\IEEEdisplaynontitleabstractindextext
%
\IEEEpeerreviewmaketitle

\IEEEraisesectionheading{\section{Introduction}\label{sec:introduction}}
\IEEEPARstart{K}{nowledge} Graph (KG), announced by Google in 2012, is a popular knowledge representation form. It focuses on entities and their relations, thus representing static knowledge. However, there is a great deal of event information in the world, which conveys dynamic and procedural knowledge. Thus, event-centric knowledge representation form like Event KG (EKG) is also essential. It has facilitated many downstream applications, such as search, question-answering, recommendation, financial quantitative investments, and text generation~\cite{app_search19,app_TCMsearch19,EventQA20,Hainan20,ERL20,EventNarrative21}.

\begin{figure*}[!htb]
	\centering
	\includegraphics[width=5.5in]{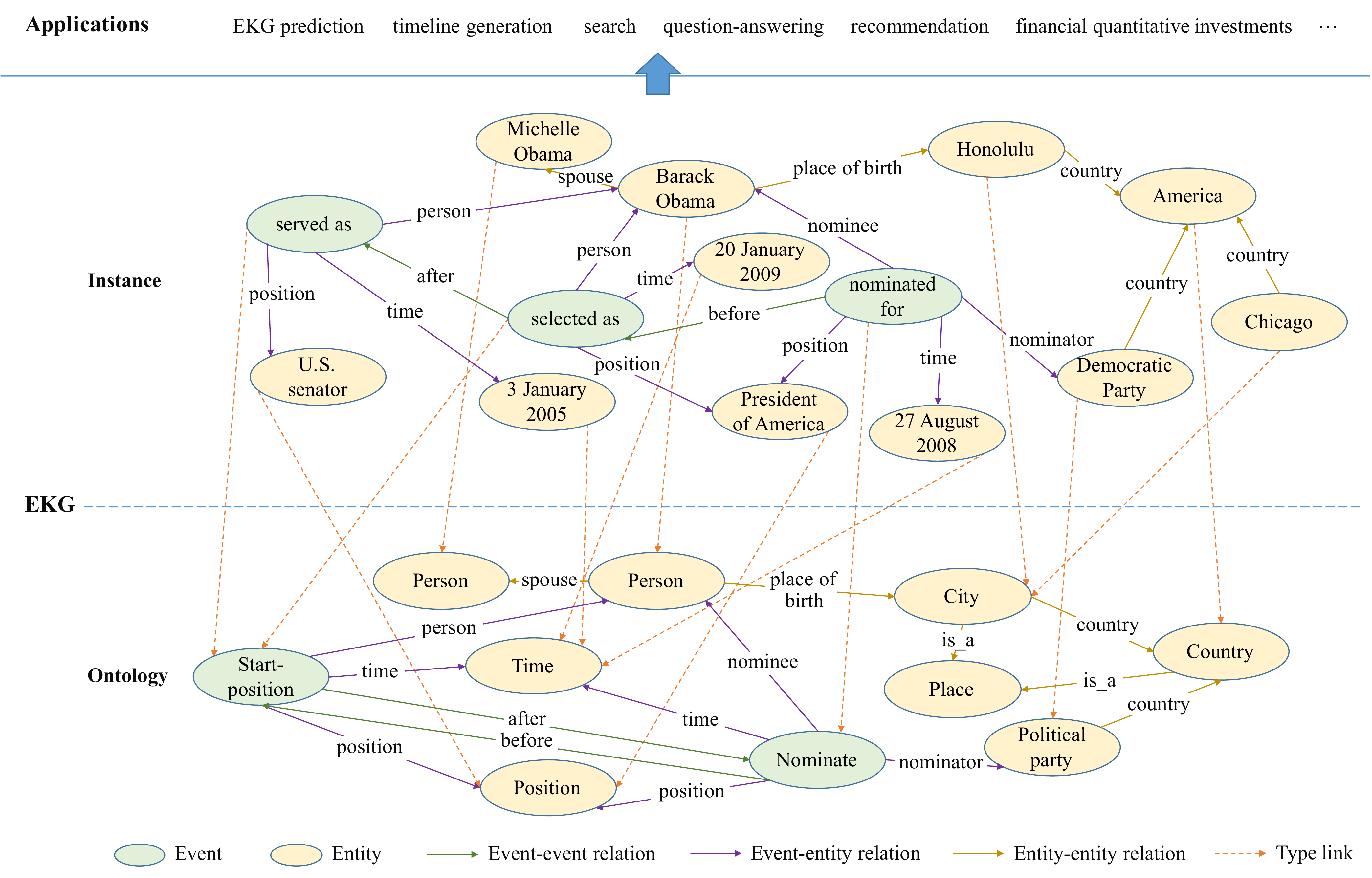}  
	\vspace{-3mm}
	\caption{A toy example of EKG and its applications.}
	\label{fig:views_relation}
	\vspace{-3mm}
	\vspace{-1mm}
\end{figure*}

This paper goes deep into what EKG is and how it develops. What do you want to know about EKG? You may be interested in how it comes into being, what is called an EKG, how to construct it, and further where it can be applied. Thus, to comprehensively introduce EKG, we see it from history, ontology, instance, and application views. As illustrated in Fig.~\ref{fig:views_relation}, ontology and instance parts cooperate to form EKG, and EKG further supports many applications. Detailedly, from the history view, we present the brief history and our derived definition of EKG (see Section~\ref{sec:history_view}). From the ontology view, we study the basic concepts related to EKG, and the tasks and methods therein, including event schema, script, and EKG schema inductions (see Section~\ref{sec:ontology_view}). From the instance view, we elaborate on event acquisition and EKG-related representative graphs/systems (see Section~\ref{sec:instance_view}). From the application view, we introduce some basic and deep applications on EKG (see Section~\ref{sec:application_view}). Moreover, the development processes and trends of the related tasks are studied thoroughly. Future directions are then pointed out in Section~\ref{sec:future_directions}, followed by conclusions in Section~\ref{sec:conclusions}.

There are also some surveys related to part of EKG, focusing on event extraction~\cite{EE_survey1,EE_survey2,EE_survey3,EE_survey4}, event modeling and mining~\cite{E_survey1}, event and event relation extractions~\cite{E_survey2}, and event coreference resolution~\cite{ECR_survey}. However, there is a lack of a comprehensive and deep survey on EKG. Actually, events are essential and non-negligible in the world. Many events happen every day, reflecting the state of the world. Thus, it is necessary to go deep into events. In this way, a comprehensive survey on EKG is of great importance.

\section{What is EKG: History View}
\label{sec:history_view}
In this section, from the history view, we present the brief history of EKG. We then derive our definition of EKG based on EKG-related concepts from the history.

\subsection{The Brief History of EKG}
\label{subsec:EKG_history}
\begin{figure*}[!htb]
	\centering
	\includegraphics[width=6.4in]{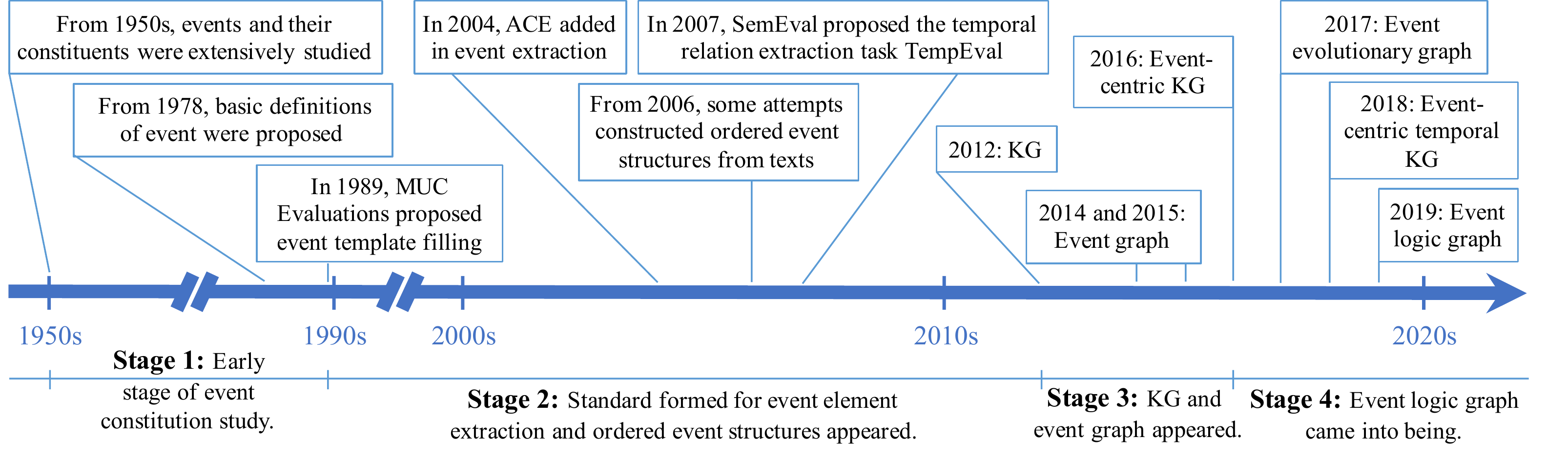}  
	\vspace{-3mm}
	\vspace{-1.5mm}
	\caption{The development history of EKG.}
	\label{fig:EKG_history}
	\vspace{-3mm}
	\vspace{-1.5mm}
\end{figure*}

EKG does not suddenly burst forth. Instead, it is the outcome of the development of Natural Language Processing (NLP) and artificial intelligence. As presented in Fig.~\ref{fig:EKG_history}, the history of EKG can be divided into four stages and dates from event, event extraction, event relation extraction, etc.

\textbf{Stage 1: Early stage of event constitution study.} From 1950s, events and their constituents were extensively studied~\cite{E_history57,E_history67,E_history78,E_history91}. For example, Davidson~\cite{E_history67} tried to get the logical form of sentences about actions. He gave an account of the logical or grammatical roles of the words in such sentences. Mourelatos~\cite{E_history78} and Pustejovsky~\cite{E_history91} explored events and proposed their basic definition. In 1978, Mourelatos~\cite{E_history78} defined events as inherently countable occurrences. In 1991, Pustejovsky~\cite{E_history91} regarded that events provide distinct representations for linguistic analysis involving the aspectual properties of verbs, adverbial scopes, argument roles, and mappings from the lexicon to syntax. 

\textbf{Stage 2: Standard formed for event element extraction and ordered event structures appeared.} In 1989, the MUC (Message Understanding Conference) evaluations proposed event template filling, initiated by Naval Ocean System Center to foster the automatic analyses of military textual messages \cite{MUC}. Given the description of events, participants were required to fill a template for each event on the page. More practically, with the unbounded information-bearing potential of the Web, the ACE (Automatic Content Extraction) program develops the capability to extract meaning therein. Starting in 2004, it added in event extraction, defined to extract event triggers and arguments, more in line with reality~\cite{ACE04}. An event trigger is a word or span that most clearly expresses the event, i.e., indicates the event type, and an argument is an entity or span that plays a specific role in the event. With the awareness of the importance of identifying the events described in a text and locating them in time, in 2007, SemEval (Semantic Evaluation) proposed the temporal relation extraction task TempEval~\cite{TempEval}. It extracts event temporal relations from texts. After that, researches on event and event relation extractions usually follow the task definitions of ACE and TempEval, respectively. Since understanding the temporal flow of discourse is significant for text comprehension, from 2006, there have been some attempts to construct ordered event structures from texts, such as temporal graph~\cite{TG06} and event timeline~\cite{ENT12,ETC12}.

\textbf{Stage 3: KG and event graph appeared.} Notably, in 2012, to enhance the results returned by Google searches, Google proposed KG with all the gathered knowledge of entities and relations in a semantic network. KG has thus caught much attention in various fields since then. However, it is about entities and their relations, i.e., static knowledge, and cannot deal with events elegantly. It somehow triggers the emergence of knowledge representation form on events and their relations. In 2014, Glavaš and Šnajder~\cite{EG14} proposed event graph that structures the information about events in texts to address the need for efficient retrieval and presentation of event-related information. In this event graph, nodes are events consisting of triggers and arguments (subject, object, time, and location), and edges are event temporal relations. In 2015, Glavaš and Šnajder~\cite{EG15} further added in event coreference relations. To describe changes in the world, in 2016, Rospocher et al.~\cite{ECKGs16} proposed event-centric KG, where nodes are events identified by URIs and entities, while edges are event-event relations, event-entity relations, and general facts about entities. The event-event relations include temporal and causal ones. Action, participant, time, and location are considered for the event-entity relations, capturing (what, who, when, and where).

\begin{table*}[htb]
	\setlength{\tabcolsep}{1em}
	\centering
	\caption{Comparison of EKG-related concepts.}
	\vspace{-1mm}
	\vspace{-2mm}
	\makebox[\columnwidth][c]{
		\begin{tabular}{c|c|c|c}
			\toprule[1pt]
			Concept &Nodes &Edges &Features\\
			\midrule[0.5pt]
			KG &Entities &Entity-entity relations &\!\!\!\!Static knowledge\!\!\!\!\\
			\midrule[0.5pt]
			\makecell[c]{Event \\graph~\cite{EG14,EG15}} &\makecell[c]{Events consisting of triggers and argu-\\\!\!\!\!ments (subject, object, time, and location)\!\!\!\!\!} &\makecell[c]{Temporal relations between events \cite{EG14}; Temporal \\and coreference relations between events \cite{EG15}}&Instance level\\
			\midrule[0.5pt]
			\makecell[c]{Event-centric \\KG~\cite{ECKGs16}} &\makecell[c]{Events identified by URIs and entities} &\makecell[c]{Event-event relations (temporal and causal relations), \\Event-entity relations (action, participant, time, \\and location), and general facts about entities} &Instance level\\
			\midrule[0.5pt]
			\makecell[c]{\!\!\!\!Event evolutionary\!\!\!\!\\ graph~\cite{EEG17}} &\makecell[c]{Events (i.e., abstract, generalized, and\\ semantically complete verb phrases)} &\makecell[c]{Temporal and causal relations between events} &Schema level\\
			\midrule[0.5pt]
			\makecell[c]{Event-centric \\temporal KG~\cite{EKG18}} &events, entities, and relations &\makecell[c]{\!\!\!\!Event-event relations (subevent, previous event, and next\!\!\!\!\\\!\!\!event relations), event-entity relations (topic, participant,\!\!\!\!\\ time, and location), and entity-entity relations} &Instance level\\
			\midrule[0.5pt]
			\makecell[c]{Event logic\\ graph~\cite{ELG19}} &\makecell[c]{\!\!\!\!Abstract, generalized, and semantically\!\!\!\!\\complete event tuples $(s, p, o)$} &\makecell[c]{Temporal, causal, conditional and hypernym-\\hyponym relations between events} &Schema level\\
			\bottomrule[1pt]
		\end{tabular}
	}
	\vspace{-3mm}
	\label{table:concept_comparison}
\end{table*}

\textbf{Stage 4: Event logic graph came into being.} Recently, with the development of many real-world applications, such as event prediction, decision making, and scenario design of dialog systems, there is a great need to understand the evolution and development of events. Thus, in 2017, Li et al.~\cite{EEG17} proposed event evolutionary graph. It is similar to event graph \cite{EG14}, but its event nodes are abstract, generalized, and semantically complete verb phrases. It further considers event causal relations and reveals evolutionary patterns and development logics of events. Then, in 2018, Gottschalk and Demidova~\cite{EKG18} proposed event-centric temporal KG, where events, entities, and relations are nodes, to facilitate semantic analyses regarding contemporary and historical events on the Web, in the news, and in social media. Events therein have topical, temporal, and geographical information, and are linked to the entities participating in the events. They also considered subevent, previous event, and next event relations, and entity-entity relations. In 2019, event evolutionary graph was derived to event logic graph~\cite{ELG19}, where nodes are abstract, generalized, and semantically complete event tuples $(s, p, o)$, $s$ is the actor/subject, $p$ is the action/predicate (i.e., event trigger), and $o$ is the object. Besides, two more event relations, conditional and hypernym-hyponym, were considered.

\textbf{Generally speaking,} there are many EKG-related concepts. As presented in Table~\ref{table:concept_comparison}, event evolutionary graph~\cite{EEG17} and event logic graph~\cite{ELG19} only focus on schema-level event knowledge. Nodes in event graph~\cite{EG14,EG15} and event logic graph are composite structures, which are difficult to handle. Moreover, these EKG-related concepts all consider specific and limited event relations and argument roles. Actually, there are many event relations. Besides, an event has its own components, each consisting of an argument and the argument role that the argument plays in the event~\cite{walker2006ace}.

\subsection{The Definition of EKG}
As introduced in Section~\ref{subsec:EKG_history}, there are some EKG-related concepts with deficiencies. We follow this line but introduce richer contents, as presented in the following.

We think that EKG, centering on events, has two types of nodes, events and entities, and three types of directed edges, indicating event-event, event-entity, and entity-entity relations. As illustrated in Fig.~\ref{fig:views_relation}, the first type of relations includes many kinds of relations between events, such as temporal, causal, conditional, thematic, etc. The second type of relations represents the arguments of events, i.e., the edges are the argument roles of the entities to the linked events. The third one describes the relations between entities, such as spouse, place of birth, country, etc. Formally,

\begin{myDef}
	EKG $G=\{(s,p,o)|\{s,p\}\in N,$ $p\in E, N=N_{evt}\cup N_{ent}, E=E_{evt\text{-}evt}\cup E_{evt\text{-}ent}\cup E_{ent\text{-}ent}\}$ is a graph of events $N_{evt}$, entities $N_{ent}$, and their relations $E$, where $E_{evt\text{-}evt}$, $E_{evt\text{-}ent}$, and $E_{ent\text{-}ent}$ are the relations between events, between events and entities, and between entities, respectively. 
	\label{def:EKG}
\vspace{-1mm}
\end{myDef}

In this way, events can easily be connected by common argument entities and vice versa. KG is thus a special case of EKG, with only entity nodes and entity-entity relations.

\begin{table*}[htb]
	\setlength{\tabcolsep}{1em}
	\centering
	\caption{Comparison of methods for event schema induction.}
	\vspace{-1mm}
	\vspace{-2mm}
	\makebox[\columnwidth][c]{
		\begin{tabular}{c|c|c|c}
			\toprule[1pt]
			Method category &Description &Strengths &Weaknesses\\
			\midrule[0.5pt]
			Supervised &\makecell[c]{Features combined with \\pattern matching or classifier} &Simple and intuitive &\makecell[c]{Hard to apply to\\ new event types} \\
			\midrule[0.5pt]
			Semi-supervised &\makecell[c]{Apply heuristic rules or vector quantized \\variational autoencoder with a few annotated seeds} &Low labor costs &Not so effective \\
			\midrule[0.5pt]
			Unsupervised &\makecell[c]{Apply rules, statistics, or graph analysis\\ (combined with representation learning)} &Efficient and widely studied &More noises\\
			\bottomrule[1pt]
		\end{tabular}
	}
	\vspace{-3mm}
	\label{table:ESI_comparison}
\end{table*}

\section{What is EKG: Ontology View}
\label{sec:ontology_view}
From the ontology view, we look into the schema and related tasks. As presented in the bottom part of Fig.~\ref{fig:views_relation}, EKG schema describes the basic concepts that form EKG, including the event types, argument roles, and event relations. The first two form event schema. As for the last one, the typical script~\cite{abelson1977scripts} organizes a set of events via some event relations, which together describe the common scenario. Before introducing EKG schema induction, let us begin with event schema and script inductions in this section.

\subsection{Event Schema Induction}
\label{subsec:event_schema_induction}
Event schema can be manually designed, such as the typical ACE event schema~\cite{ACE04,walker2006ace} and FrameNet frame~\cite{baker1998berkeley}. Since manually designed event schema shows low coverage and difficulty in domain adaptation, researchers have paid attention to event schema induction. It automatically extracts event types and their argument roles from texts. Formally, 
\begin{myDef}
	\textbf{Event schema induction:} Given a set of texts $\{T_0, T_1, ..., T_{l}\}$, it identifies the event schema, including all event types $\{{tp}_0, ..., {tp}_{\tau}\}$ and all argument roles $\{{rl}_0^i, ..., {rl}_{\rho}^i\}$ for each event type ${tp}_i$ ($0\leq i \leq \tau$). 
	\label{def:ESI}
\vspace{-1mm}
\end{myDef}
For example,
\begin{myExam}
	Input: $T_0$: Barack Obama previously served as a U.S. senator from 3 January 2005. Then, he was selected as the President of America on 20 January 2009.\\
    $T_1$: Before winning the presidential election, Obama was nominated for the President of America on 27 August 2008.\\
	Output: ${tp}_0$: Start-position, ${rl}_0^0$: person, ${rl}_1^0$: position, ${rl}_2^0$: time; ${tp}_1$: Nominate, ${rl}_0^1$: nominee, ${rl}_1^1$: position, ${rl}_2^1$: time.
	\label{exam:ESI}
\vspace{-1mm}
\end{myExam}

Existing methods for this task can be divided into supervised, semi-supervised, and unsupervised ones.

\textbf{Supervised methods} are applied in early studies. They learn from annotated data and then induce event schema from new texts \cite{lehnert-etal-1992-university,chinchor-etal-1993-evaluating,freitag-1998-toward,chieu-etal-2003-closing}. For example, methods in the third MUC evaluations used pattern matching (e.g., regular expressions), syntax-driven techniques which combine syntactic analyses with semantics and subsequent processing, or integrated syntax-driven techniques into pattern matching for event schema induction~\cite{chinchor-etal-1993-evaluating}. Chieu et al.~\cite{chieu-etal-2003-closing} adopted semantic and discourse features, and built a classifier, e.g., maximum entropy, Support Vector Machine (SVM), naive Bayes, or decision tree, for each argument role.

\textbf{Semi-supervised methods} start with a few annotated seeds to induce event schema \cite{yangarber-etal-2000-automatic,surdeanu-etal-2006-hybrid,patwardhan-riloff-2007-effective,huang-ji-2020-semi}. For example, Patwardhan and Riloff~\cite{patwardhan-riloff-2007-effective} created a self-trained SVM to identify relevant sentences for the domain of interest and then extracted domain-relevant event schema via semantic affinity. The self-training begins with seed patterns and relevant and irrelevant documents. The following event schema extraction is based on heuristic rules upon syntactic analyses. The extracted results were ranked by semantic affinity based on frequency to keep the top results. Huang and Ji~\cite{huang-ji-2020-semi} discovered unseen event types by leveraging annotations available for a few seen types. They designed a vector quantized variational autoencoder to learn an embedding for each seen or unseen event type and optimized it using seen event types. A variational autoencoder was further introduced to enforce the reconstruction of each event trigger conditioned on its event type distribution.

\textbf{Unsupervised methods} remove the requirements of annotated data and are widely applied \cite{sudo-etal-2003-improved,filatova-etal-2006-automatic,chambers-jurafsky-2011-template,balasubramanian-etal-2013-generating,ESI13,ESI15,huang2016liberal,OSEP18,qasemizadeh-etal-2019-semeval,yamada-etal-2021-semantic,shen-etal-2021-corpus}. For example, Chambers and Jurafsky~\cite{chambers-jurafsky-2011-template} viewed event schema induction as discovering unrestricted relations. They used Pointwise Mutual Information (PMI) to measure the distances between events and clustered events according to the distances. Then, they induced the argument roles of events via the syntactic relations. Balasubramanian et al.~\cite{balasubramanian-etal-2013-generating} used co-occurrence statistics of $(s, p, o)$ pairs to build a graph with these triples as nodes and edges were weighted by the symmetric conditional probabilities of the involved triple pairs. Triples therein were normalized using stemmed headwords and semantic types. They started with high-connectivity nodes as seeds. Then, they applied graph analysis to find closely related triples to the seeds and merged their argument roles to create event schema. Chambers~\cite{ESI13} proposed the first generative model for schema induction similar to LDA~\cite{LDA}. Nguyen et al.~\cite{ESI15} further introduced entity disambiguation.

More recent studies introduced representation learning to induce event schema unsupervisedly \cite{huang2016liberal,OSEP18,qasemizadeh-etal-2019-semeval,yamada-etal-2021-semantic,shen-etal-2021-corpus}. For example, Yuan et al.~\cite{OSEP18} proposed a two-step framework. They first detected event types by clustering news articles. Then, they proposed a graph-based model that exploits entity co-occurrence to learn entity embeddings and clustered these embeddings into argument roles. Methods in the 2019 International Workshop on Semantic Evaluation applied pretrained language models, such as BERT~\cite{devlin2018bert}, to get word embeddings \cite{qasemizadeh-etal-2019-semeval}. They then clustered these embeddings with hand-crafted features and aligned them to the event types and argument roles of the existing event schema (e.g., FrameNet). Yamada et al.~\cite{yamada-etal-2021-semantic} thought previous studies focused too much on the superficial information of verbal event triggers and proposed to use masked word embeddings from BERT to get deep contextualized word embeddings. They then applied a two-step clustering method, which clusters instances of the same verb according to the embeddings and further clusters across verbs. Finally, each generated cluster was regarded as an induced schema.

\textbf{In a word,} as presented in Table~\ref{table:ESI_comparison}, for supervised methods, they are hard to apply to new event types, which limits their usage. For semi-supervised and unsupervised methods, automatically derived event schema is noisy and difficult to be aligned. So far, these techniques are still not so applicable to building the event schema for an EKG.

\subsection{Script Induction}
A script can be seen as a stereotypical structure of event schema that expresses a specific scenario. Specifically, it organizes a set of events into a certain structure (usually according to their temporal relations). The ``events'' in scripts, called script events, are event schema rather than instances.

\begin{figure}
	\centering
	\includegraphics[width=3.3in]{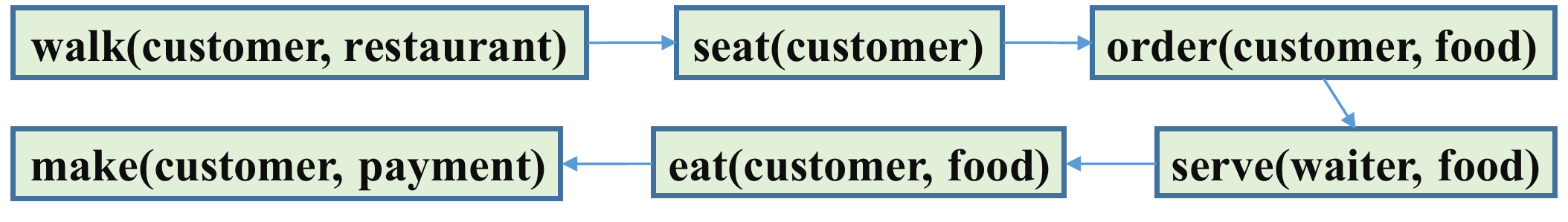}
	\vspace{-3mm}
	\caption{A typical example of the induced restaurant script.}
	\vspace{-3mm}
	\vspace{-2mm}
	\label{fig:script}
\end{figure}

In early studies, scripts were manually designed~\cite{abelson1977scripts,raskin-etal-2003-genesis}. In more recent studies, researchers tried to extract script automatically from texts, i.e., script induction. Formally, 
\begin{myDef}
	\textbf{Script induction:} Given a set of texts $\{T_0, T_1, ..., T_{l}\}$, it identifies all the scripts $\{{\Phi}_0, ..., {\Phi}_\epsilon\}$, and each script ${\Phi}_i$ ($0\leq i \leq \epsilon$) is a coherent event schema sequence $\{{tp}_0^i({rl}_0^{i,0}, ..., {rl}_{n_0^i}^{i,0}), ..., {tp}_\phi^i({rl}_0^{i,\phi}, ..., {rl}_{n_\phi^i}^{i,\phi})\}$, where $n_j^i$ ($0\leq j \leq \phi$) is the number of argument roles of event type ${tp}_j^i$, ${rl}_\theta^{i,j}$ is the $\theta$-th argument role of ${tp}_j^i$. 
	\label{def:SI}
\vspace{-1mm}
\end{myDef}
Fig.~\ref{fig:script} is a typical example of the induced restaurant script, modeling the scenario of a customer eating food in a restaurant.

The main challenge is to learn the scenario-specific argument roles automatically. Chambers and Jurafsky~\cite{chambers-jurafsky-2009-unsupervised} used the most frequent headword in an entity coreferential chain as the argument role of this entity. Regneri et al.~\cite{regneri-etal-2010-learning} collected descriptions of script-specific event sequences from volunteers. They then computed a graph representation of the script’s temporal structure. This graph makes statements about what phrases describe events of a scenario and in what order they occur. Cheung et al.~\cite{PFI13} defined a joint distribution over the words in a document and their script assignments. Specifically, they combined a script Hidden Markov Model (HMM) \cite{HMM} with an event HMM, where the first HMM models script transition and emits events, and the second one models event transition within a script and emits argument roles. Orr et al.~\cite{10.5555/2892753.2892770} also learned scripts based on HMM. Differently, the states of the HMM are event types in scripts, and observations are sentences describing the event instances, which were obtained via a clustering algorithm. Then, they started with a fully enumerated HMM of the event sequences and merged states and deleted edges (state transitions) to improve the posterior probability of the structure and parameters given the data. Weber et al.~\cite{weber-etal-2020-causal} argued that correlation-based approaches are insufficient and induced script based on the causal effect between events, defined via interventions using a Bayesian network-based method. Weber et al.~\cite{weber2021schema} and Ciosici et al.~\cite{ciosici-etal-2021-machine} further induced script via human-machine collaboration manner.

\textbf{In general,} among these script induction methods, the simple one proposed by Chambers and Jurafsky~\cite{chambers-jurafsky-2009-unsupervised} is usually used in the studies of script application \cite{ERL16_2,wang-etal-2017-integrating,ijcai2018-584}. However, it is far from satisfactory. For example, suppose a coreferential chain contains ``Jobs'', ``Steve Jobs'', and ``CEO Jobs''. The above method will use ``Jobs'' instead of ``CEO'' as the argument role, which is too specific and lacks generalization. It is still a problem to learn high-quality scenario-specific argument roles effectively and efficiently.

Script can be seen as rules about events, forming the evolution patterns of events in a certain scenario. Its one basic application is to predict what will happen in the future via script event prediction. Specifically, the known real-world events are generalized into script events. Then, they are used to derive the subsequent script events, called script event prediction. Finally, the predicted subsequent script events can be instantiated to real-world events. We will describe more details of script event prediction in section~\ref{subsubsec:script_event_prediction}.

\subsection{EKG Schema Induction}
There are few studies for EKG schema induction. Formally, 
\begin{myDef}
	\textbf{EKG schema induction:} Given a set of EKGs $\{G_0, ..., G_\xi\}$, it identifies the schema $\mathbb{G} = \{\mathbb{E}, \mathbb{R}_{evt\text{-}evt}, \mathbb{R}_{ent\text{-}ent}\}$, where $\mathbb{E}$ is the event schema, including event types and their argument roles constraining the arguments (i.e., event type-entity type relations), and $\mathbb{R}_{evt\text{-}evt}$ and $\mathbb{R}_{ent\text{-}ent}$ are the sets of relations between event types and between entity types, respectively.
	\label{def:EKGSI}
\vspace{-1mm}
\end{myDef}
Take Fig.~\ref{fig:views_relation} for a simple example, with the instance graph as input, EKG schema induction outputs the schema (see the bottom part of Fig.~\ref{fig:views_relation}).

Since EKG-related studies have not been put forward for a long time, researches on EKG schema induction are much scarcer and newer. In 2020, Li et al.~\cite{EGSI20} first studied event graph schema induction, focusing on rich event components and event-event connections. A path language model was proposed to build an event graph schema repository, where two event types are connected through event-event paths involving entities that fill important argument roles. These entities were replaced by their types. However, this work only pays attention to the connections between events. In 2021, Li et al.~\cite{ESI21} further focused on all three types of relations, i.e., event-event, event-entity, and entity-entity relations. They regarded schema as hidden knowledge to guide the generation of event graphs and learned via maximizing the probability of these instance graphs. However, for event-event relations, only temporal relations were considered.

\textbf{Generally speaking,} existing few studies on EKG schema induction consider limited relation types. Thus, there is a long way for EKG schema induction to induce overall schema for EKG.

\section{What is EKG: Instance View}
\label{sec:instance_view}
From the instance view, this section introduces how to construct an EKG, i.e., event acquisition and EKG-related representative graphs/systems.

\subsection{Event Acquisition}
\begin{figure}[!htb]
	\centering
	\vspace{-2mm}
	\includegraphics[width=3.3in]{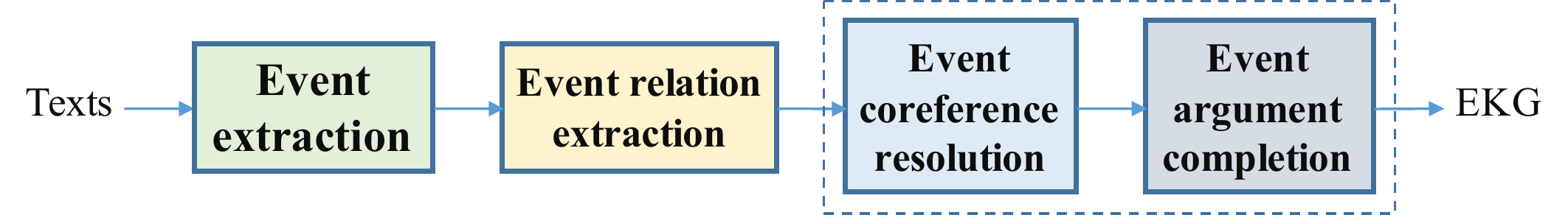}
	\vspace{-3mm}
	\caption{Different steps in event acquisition.}
	\label{fig:EKG_construction}
	\vspace{-3mm}
\end{figure}

Event acquisition is essential for EKG construction. As presented in Fig.~\ref{fig:EKG_construction}, it mainly includes event extraction, event relation extraction, event coreference resolution, and event argument completion. The former two tasks are basic, and the latter two are important for constructing a better EKG. Due to the complex event structures, they are more challenging than entity extraction, entity relation extraction, and entity coreference resolution in KG construction. In this section, we review their development and future trends.

\subsubsection{Event Extraction}
As the primary step in constructing EKG, event extraction is to extract structured event information from texts, including event triggers with types and arguments with roles. Thus, there are two subtasks, trigger detection and argument extraction. The former identifies the event trigger words and assigns them with predefined types or clustered classes, while the latter identifies the arguments and assigns them with argument roles of the triggered events. Formally,
\begin{myDef}
\textbf{Event extraction:} Given a text $S = \{w_0, w_1, ..., w_n\}$ of $n$ words, it identifies all event triggers $\{{trg}_0, ..., {trg}_m\}$ and predicts the event type ${tp}_i$ for each event trigger ${trg}_i = \{w_{b_i^{trg}}, ..., w_{d_i^{trg}}\}$, where $0\leq i \leq m$, $b_i^{trg}$ and $d_i^{trg}$ are the beginning and end indexes of ${trg}_i$. It also identifies event arguments $\{{arg}_0^i, ..., {arg}_k^i$\} for ${trg}_i$ and predicts the argument role ${rl}_j^i$ for each event argument ${arg}_j^i = \{w_{b_{ij}^{arg}}, ..., w_{d_{ij}^{arg}}\}$, where $0\leq j \leq k$, $b_{ij}^{arg}$ and $d_{ij}^{arg}$ are the beginning and end indexes of ${arg}_j^i$.
	\label{def:EE}
\vspace{-5mm}
\end{myDef}
For example,
\begin{myExam}
    Input: $S$: Barack Obama previously served as a U.S. senator from 3 January 2005.\\
    Output: \\
    ${trg}_0$: served as~~~~~~~~~~~~~~~ ${tp}_0$: Start-position;\\
    ${arg}_0^0$: Barack Obama~~~~~~ ${rl}_0^0$: person;\\
    ${arg}_1^0$: U.S. senator~~~~~~~~~~ ${rl}_1^0$: position;\\
    ${arg}_2^0$: 3 January 2005~~~~~~ ${rl}_2^0$: time.
	\label{exam:EE}
\vspace{-1mm}
\end{myExam}

\begin{figure}[!htb]
	\centering
	\vspace{-2mm}
	\includegraphics[width=3.3in]{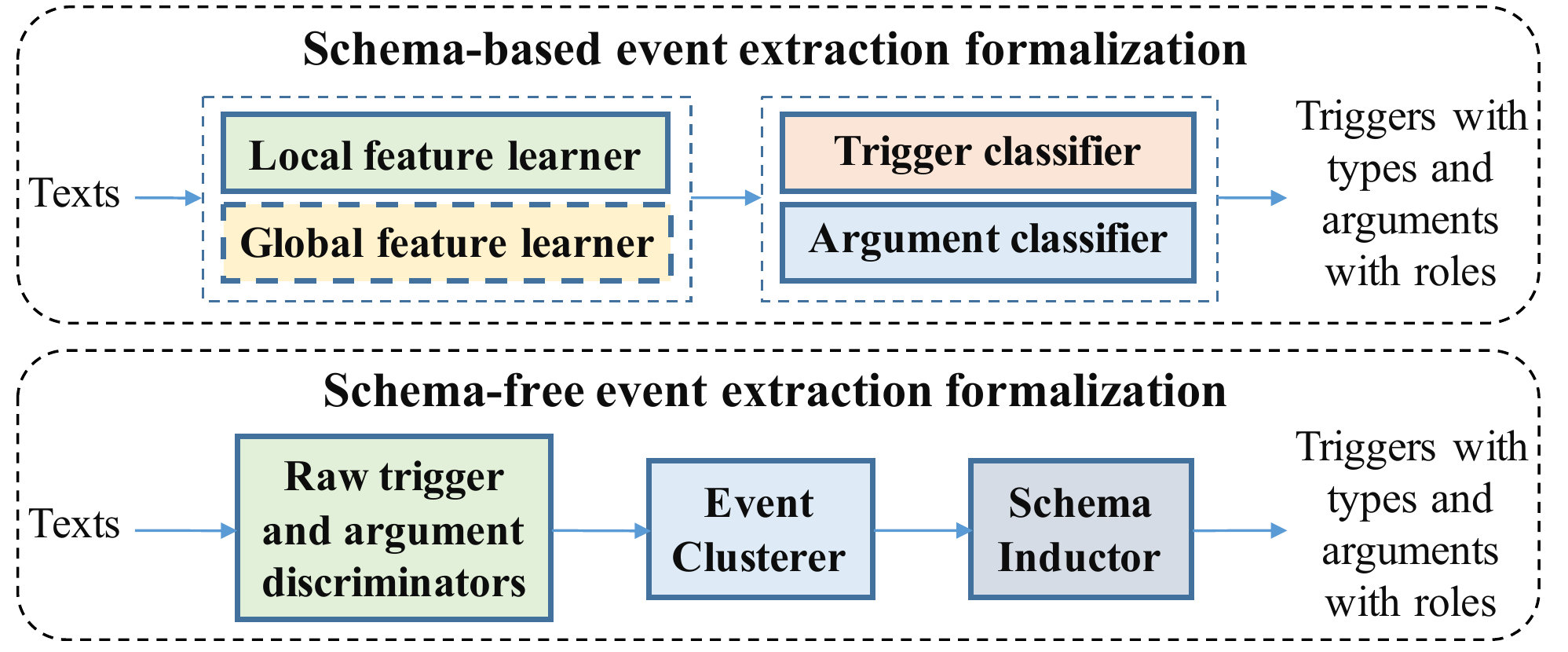}
	\vspace{-3mm}
	\caption{Existing formalizations of event extraction.}
	\label{fig:EE_formalization}
	\vspace{-2mm}
\end{figure}

Event extraction is divided into schema-based and schema-free ones concerning if there is a predefined schema. As illustrated in the upper part of Fig.~\ref{fig:EE_formalization}, existing schema-based methods pass texts to the feature learners to obtain local (and global) features. Upon them, the classifiers of triggers and arguments output the probability distributions on the predefined schema and get the answers based on the peaks. As for schema-free event extraction formalization (the bottom part of Fig.~\ref{fig:EE_formalization}), texts are passed to the discriminators to get raw triggers and arguments, which are clustered into groups to induce the event schema and get the answers. Simple unsupervised event schema induction methods (see Section~\ref{subsec:event_schema_induction}) are usually used therein. Specifically, considering the input scale, schema-based event extraction is further grouped into sentence- and document-level ones, and schema-free event extraction is also called open-domain one. Compared to document-level and open-domain event extractions, sentence-level one is more extensively studied.

\textbf{Sentence-level event extraction} extracts event triggers and arguments in a sentence. Early approaches designed elaborate features and applied statistical learning methods \cite{ahn2006stages,chen-ji-2009-language,liao2010using}. Recently, with the construction of large-scale datasets (e.g., ACE~\cite{walker2006ace}, TAC-KBP~\cite{song2015light}, and RAMS~\cite{ebner2019multi}) and the development of deep learning, researchers have adopted neural networks to extract features automatically. According to the feature scopes, they are divided into three categories:

\textit{Explore intra-subtasks features.} Most methods employ a Convolutional Neural Network (CNN) or Recurrent Neural Network (RNN) to extract the intra-subtasks features and follow a pipeline framework performing each event extraction subtask sequentially. Chen et al.~\cite{chen2015event} proposed dynamic multi-pooling CNN to extract the word and various sentence-level features for candidate words. However, as a CNN-based method, this work cannot properly handle the sequential relations and long-range dependencies among the words in a sentence. Thus, Chen et al.~\cite{10.1007/978-3-319-47674-2_17} adopted bidirectional dynamic multi-pooling Long Short-Term Memory network (LSTM). Besides, they designed a tensor layer to explore interactions between candidate arguments. With the development of the pretrained language models like BERT and ELMo~\cite{peters2018deep}, recent researchers have introduced them into event extraction. For example, Yang et al.~\cite{yang2019exploring} added a multi-classifier on the BERT for trigger detection and multiple sets of binary classifiers on the BERT for argument extraction. Deng et al.~\cite{deng2021ontoed} enriched event schema with event-event relations, such as temporal, causal, and hierarchical relations, to improve trigger detection based on BERT. Recently, Du and Cardie~\cite{du2020event} and Liu et al.~\cite{liu2020event} proposed machine reading comprehension frameworks upon BERT, which adopt question answering to extract events.  

\textit{Explore inter-subtasks features.} These methods explore the inter-subtasks features of trigger detection and argument extraction, and usually follow a joint framework, easing error propagation of the above pipeline framework. Nguyen et al.~\cite{nguyen2016joint} proposed a joint model based on bidirectional gated recurrent units and Fully Connected Network (FCN). They introduced memory vectors to store the dependencies among trigger subtypes and argument roles. To further make use of syntactic features, Sha et al.~\cite{sha2018jointly} added dependency bridges to connect syntactically related words in RNN and built a tensor layer on each pair of candidate arguments to capture intensive argument-level interaction. To capture the long-range dependencies more efficiently, Liu et al.~\cite{liu2018jointly} used an attention-based Graph Convolution Network (GCN) to aggregate the word information through the paths of the syntactic tree of the sentence and capture more interactions among the candidate triggers and arguments.

\begin{table*}[htb]
	\setlength{\tabcolsep}{1em}
	\centering
	\caption{Comparison of methods for event extraction.}
	\vspace{-1mm}
	\vspace{-2mm}
	\makebox[\columnwidth][c]{
		\begin{tabular}{c|c|c|c}
			\toprule[1pt]
			Method category & Description & Strengths & Weaknesses\\
			\midrule[0.5pt]
			\makecell[c]{Sentence-level event extraction: \\Explore intra-subtasks features} & \makecell[c]{Follow a pipeline framework \\performing each subtask sequentially} &\makecell[c]{Intuitive and with \\low model complexity} &Error propagation\\
			\midrule[0.5pt]
			\makecell[c]{Sentence-level event extraction: \\Explore inter-subtasks features} & \makecell[c]{Follow a joint framework performing\\ each subtask simultaneously} &\makecell[c]{Capture the dependencies\\ among event elements} & High model complexity \\
			\midrule[0.5pt]
			\makecell[c]{Sentence-level event extraction: \\Explore inter-IE features} &\makecell[c]{Adopt a multi-task learning framework \\introducing other related IE tasks} &\makecell[c]{Benefit from more global \\features and is more effective} &High model complexity\\
			\midrule[0.5pt]
			Document-level event extraction &\makecell[c]{Extract events in multiple sentences \\via the local and global features} &Fit real-world scenarios &\makecell[c]{Less studied and\\ far from solved}\\
			\midrule[0.5pt]
			Open-domain event extraction &\makecell[c]{Extract simple event elements through \\clustering without a predefined schema} &Need no annotation data &Less effective \\
			\bottomrule[1pt]
		\end{tabular}
	}
	\vspace{-3mm}
	\label{table:EE_comparison}
\end{table*}

\textit{Explore inter-IE features.} To better model the semantic information of event elements, recent researchers have explored more global features via introducing other related Information Extraction (IE) tasks. With a multi-task learning framework, these IE tasks benefit from each other. Nguyen and Nguyen~\cite{nguyen2019one} employed a bidirectional RNN to learn word embeddings, over which entity and event extractions were conducted via the classifiers based on FCN and softmax. Wadden et al.~\cite{wadden2019entity} further handled entity extraction, entity relation extraction, event extraction, and coreference resolution. After encoding sentences via BERT, they enumerated text spans and constructed a span graph. Span embeddings were updated by integrating embeddings from their neighbors and passed to the FCN-based classifiers of all the tasks. However, these studies handle the tasks separately. To better explore the inter-IE features, Lin et al.~\cite{lin2020joint} extracted entities, entity relations, and events simultaneously. They computed local scores for all the candidate triggers, entities, and their pairwise links via the BERT embeddings and FCN-based classifiers. Then, they searched for the globally optimal results with a beam search-based decoder incorporating inter-dependencies among the candidate triggers and entities.

\textbf{Document-level event extraction.} Sentence-level event extraction assumes that event trigger and its arguments are in the same sentence. However, they usually scatter across multiple sentences in a document in real-world scenarios. Thus, document-level event extraction is practical. It is also more challenging, since arguments may exist in different sentences and a document usually contains multiple events.

Early approaches used handcrafted features to model events and involved entities. Then, they extracted events in the documents via statistical learning methods \cite{ji2008refining,berant2014modeling,yang2016joint}. These features and annotated data are usually expensive to obtain. To address this problem, Yang et al.~\cite{yang2018dcfee} used a distantly supervised method to label events in the documents automatically. With the annotated data, they trained a sentence-level event extraction model with a Bidirectional LSTM (BiLSTM) and a Conditional Random Field (CRF) layers. Based on the extracted events, they padded the missing arguments. This framework tackles argument scattering in a pipeline way. To handle document-level event extraction end to end, Zheng et al.~\cite{zheng2019doc2edag} transformed events into an entity-based directed acyclic graph. Then, they transformed event extraction into several sequential path-expanding subtasks. Xu et al.~\cite{xu2021document} also did this transformation but explored more interactions among sentences and events via applying Graph Neural Network (GNN) on the graph of sentences and entities and storing the extracted events into a global memory to facilitate the current extraction. Differently, Lou et al.~\cite{lou2021mlbinet} proposed a multi-layer bidirectional network to capture document-level information and event inter-dependency simultaneously for trigger detection.

\textbf{Open-domain event extraction.} Unlike sentence- and document-level event extractions, open-domain one has no predefined schema. It extracts events mainly from long texts like newswires and short ones like social media streams.

For long texts, Rusu et al.~\cite{rusu2014unsupervised} treated verbs as event triggers and analyzed the dependency paths between verbs and other syntactic elements (entities, time expressions, subjects, and objects) to identify arguments. However, syntactic relations are too simple to describe complex events. To extract events and induce schema simultaneously, Huang et al.~\cite{huang2016liberal} identified all the nouns and verbs that can be matched to the existing schema like FrameNet as candidate event triggers and identified their candidate arguments via manually-selected semantic relations. Then, they were clustered based on their embeddings from a tensor-based model and named to obtain the schema and extracted results via the mappings to the existing schema. Differently, Liu et al.~\cite{liu2019open} learned the joint probability distribution of a news cluster with headwords, contextual features, and latent event type. Then, they adopted the learned distributions to cluster news and applied a series of rules to get the final results.

Unlike long texts that need inducing a complex schema, the main event components in social media streams these short texts are entities, time, location, and keywords. Abdelhaq et al.~\cite{abdelhaq2013eventweet} detected events important in a small area from a tweet stream. They extracted event keywords based on frequency. Then, they calculated the spatial density distributions over the usage ratios of the keywords at particular locations. The ones with small entropy (occur at a few locations) were clustered by their distributions to get the events. To extract more information, Wang et al.~\cite{wang2015seeft} fused Twitter and related Web pages to extract events, their times, locations, and titles by tweet- and page-based CRFs, respectively. However, multiple mentions may refer to the same entity and would be wrongly assigned to different events by the above methods. Thus, Zhou et al.~\cite{zhou2017event} proposed a non-parametric Bayesian mixture model for event extraction from Twitter, which applies word embeddings to deal with this issue. Xu et al.~\cite{xu2019jointly} further used a BiLSTM, a control gate, and a CRF layers to extract events from Twitter.

\textbf{In a word,} although event extraction has been studied for a long time (see Table~\ref{table:EE_comparison}), the performance is unsatisfactory, especially for argument extraction. Thus, it may introduce noises into EKG construction. Moreover, partly due to the low performance of argument extraction, some researches on EKG applications like script event and temporal KG predictions (see Sections~\ref{subsubsec:script_event_prediction} and \ref{subsubsec:TKG_prediction}) only consider the simpler fixed argument roles, subject, object, and indirect object or time, instead of well-defined complex schema. Thus, improving the performance of this primary event extraction task is of great importance.

\subsubsection{Event Relation Extraction}
Besides event extraction, event relation extraction is fundamental to EKG construction. It extracts relations between events from texts and links events to get an EKG. Formally, 
\begin{myDef}
\textbf{Event relation extraction:} Given a text $S = \{w_0, w_1, ..., w_n\}$ of $n$ words, where a pair of events are specified as $(e_1, e_2)$, it identifies their relation $r(e_1, e_2)$.
	\label{def:ERE}
\vspace{-1mm}
\end{myDef}
Currently, events are simplified as verbs. For example, 
\begin{myExam}
    Input: $S$: Barack Obama previously $\langle e_1 \rangle $ served as $\langle / e_1 \rangle $ a U.S. senator from 3 January 2005. Then, he was $\langle e_2 \rangle $ selected as $\langle / e_2 \rangle $ the President of America on 20 January 2009.\\
    Output: $r(e_1, e_2)$: before.
	\label{exam:EE}
	\vspace{-1mm}
\end{myExam}

\begin{figure}[!htb]
	\centering
	\vspace{-4mm}
	\includegraphics[width=2.6in]{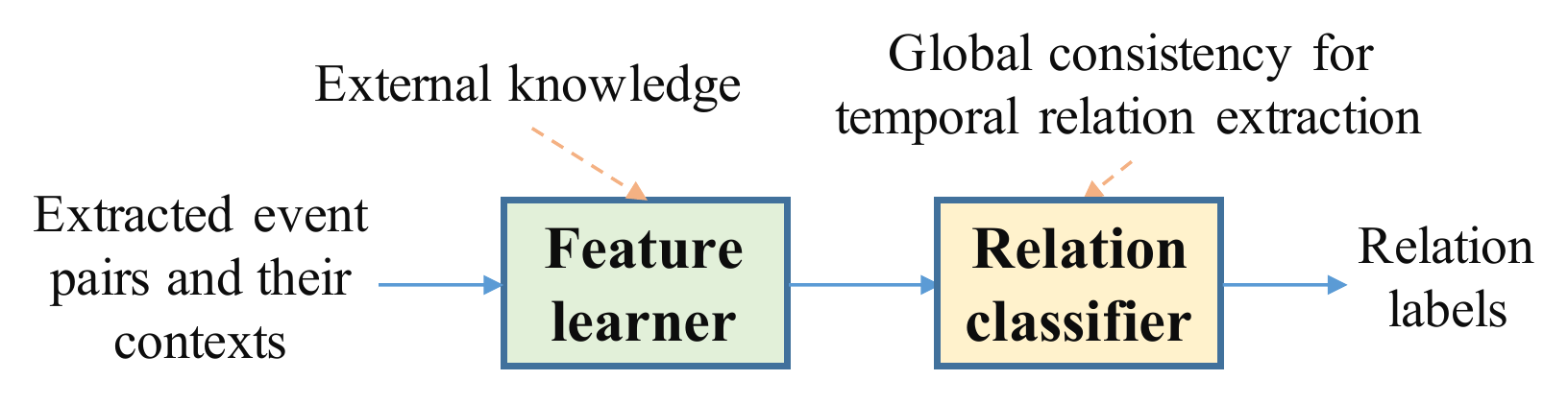}
	\vspace{-3mm}
	\vspace{-1mm}
	\caption{Existing formalization of event relation extraction.}
	\label{fig:ERE_formalization}
	\vspace{-1mm}
	\vspace{-1mm}
\end{figure}

The main event relations are event temporal and causal relations. For short, we use the terms temporal and causal relations, respectively. The former describes the temporal order between events. The latter describes the causality between events and is a subset of temporal relation. These two types of relation extractions share similar research lines and are usually formalized as text classification given the event pairs and contexts. As presented in Fig.~\ref{fig:ERE_formalization}, extracted event pairs and their contexts are passed to the feature learner to capture helpful information. Based on these features, the relation classifier outputs the relation labels. Some methods also introduce external knowledge. For temporal relation extraction, some methods consider the important global consistency problem. For example, if the classifier gets the results: A $before$ B, B $before$ C, and C $before$ A, then there is a conflict. Actually, the first two imply A $before$ C.

\textbf{Early methods relied on the manually designed syntactic and semantic features~\cite{chang-choi-2004-causal,mani2006machine,chambers-etal-2007-classifying,blanco2008causal,rink-etal-2010-learning,rink-etal-2010-learning,zhao-etal-2016-event}.} They employed machine learning models, such as naive Bayes, maximum entropy, or SVM as the classifier to recognize relations. For example, Chambers et al.~\cite{chambers-etal-2007-classifying} applied SVM as the classifier for temporal relation extraction. Features therein include temporal attributes of events, such as tense, event type, modality, polarity, and other linguistic features. Rink et al.~\cite{rink-etal-2010-learning} used graph patterns as features to train an SVM classifier. Specifically, they built a graph representation for the sentence of events, which encodes lexical, syntactic, and semantic information. They automatically extracted graph patterns from such graph representations, sorted by the relevance in determining the causality between events. Zhao et al.~\cite{zhao-etal-2016-event} proposed a restricted hidden naive Bayes model to handle feature interactions for causal relation extraction. Besides contextual, syntactic, and position features, they utilized a new feature of causal connectives, obtained from the similarities of the syntactic dependency structures of sentences expressing causality. 

The above methods for temporal relation extraction only focus on pairwise decisions. Thus, many methods further considered global consistency \cite{TG06,chambers-jurafsky-2008-jointly,yoshikawa-etal-2009-jointly,ETC12,chambers-etal-2014-dense,mirza2016catena,ning2017structured}. Some applied Integer Linear Programming (ILP) on top of the above local classifiers~\cite{TG06,chambers-jurafsky-2008-jointly,ETC12}. Chambers et al.~\cite{chambers-etal-2014-dense} proposed a cascade architecture with a sequence of classifiers ordered by their precision. The classifiers were run in order starting with the most precise one. Global consistency was enforced by inferring all the transitive relations from the results of earlier classifiers before passing them to the next classifier. Mirza and Tonelli~\cite{mirza2016catena} further applied similar methods as temporal and causal relation extractors. Different from these pipeline manners, some methods considered global consistency in the learning stage~\cite{yoshikawa-etal-2009-jointly,ning2017structured}. Yoshikawa et al.~\cite{yoshikawa-etal-2009-jointly} proposed a Markov logic model and captured global consistency via the addition of weighted first-order logic formulae. Ning et al.~\cite{ning2017structured} trained the local classifier with feedback by performing global inference in each round of the learning.

Besides the above intra-sentence causal relation extraction, Gao et al.~\cite{gao-etal-2019-modeling} designed lexical, potential causal, and syntactic features for document-level one. These global and fine-grained aspects of causality were learned via ILP.
\begin{table*}[htb]
	\setlength{\tabcolsep}{1em}
	\centering
	\caption{Comparison of methods for event relation extraction.}
	\vspace{-1mm}
	\vspace{-2mm}
	\makebox[\columnwidth][c]{
		\begin{tabular}{c|c|c|c}
			\toprule[1pt]
			Method category &Description &Strengths &Weaknesses\\
			\midrule[0.5pt]
			Feature-based &\makecell[c]{Use manually designed \\syntactic and semantic features} &High precision &\makecell[c]{Difficult to transfer\\ to other domains}\\
			\midrule[0.5pt]
			Neural network-based &\makecell[c]{Automatically learn features\\ via neural networks} &Easy to transfer & Require large data\\
			\midrule[0.5pt]
			Pretrained language model-based &\makecell[c]{Adopt pretrained language models\\ to encode event sentences} &\makecell[c]{Easy to transfer and utilize external\\ knowledge from large corpus} &Heavy computation\\
			\bottomrule[1pt]
		\end{tabular}
	}
	\vspace{-3mm}
	\label{table:ere_comparison}
\end{table*}

\textbf{More recent methods utilized neural networks to learn useful features for extraction automatically~\cite{dligach2017neural,tourille-etal-2017-neural,dasgupta2018automatic,han-etal-2019-deep,ning2019improved,li-mao-2019-knowledge}.} They applied CNN or LSTM to encode event sentences, followed by a relation classifier based on FCN and softmax. To further make globally consistent decisions for temporal relation extraction, Han et al.~\cite{han-etal-2019-deep} adopted an SVM-based algorithm incorporating transitivity constraints, and Ning et al.~\cite{ning2019improved} employed ILP. Differently, Cheng and Miyao~\cite{cheng-miyao-2017-classifying} adopted BiLSTM along dependency paths of event sentences for temporal relation extraction.

To further make use of external knowledge, Ning et al.~\cite{ning2019improved} applied a Siamese network to a temporal common sense knowledge base, whose output was concatenated with the LSTM output of event sentences, for temporal relation extraction. Li and Mao~\cite{li-mao-2019-knowledge} proposed a knowledge-oriented CNN for causal relation extraction, where the filters were generated from lexical knowledge bases to represent causal keywords and cue phrases. They also combined a conventional CNN to learn other features of causal relations.

To further introduce other related tasks, Han et al.~\cite{han-etal-2019-joint} extracted event and temporal relation jointly to avoid error propagation in the pipeline manners, which extract them sequentially. Specifically, after encoding sentences via BiLSTM, they computed the probability of being an event and the softmax distribution over all the temporal relation labels. Global consistency was considered via the last SVM-based layer incorporating constraints of transitivity, symmetry, etc. Wang et al.~\cite{wang2020joint} proposed a similar method for temporal and subevent relation extractions but replaced the SVM-based layer with a differentiable constrained learning layer. The constraints were converted into differentiable functions. Different from these hard constraints, Han et al.~\cite{han-etal-2020-domain} improved similar networks of event and temporal relation extractions by introducing corpus statistics as soft constraints.

\textbf{Notably, since pretrained language models perform well on many NLP tasks, researchers have introduced them, such as BERT, into event relation extraction.} Many neural network-based researches simply used their pretrained word embeddings~\cite{han-etal-2019-deep,ning2019improved,han-etal-2019-joint,wang2020joint,han-etal-2020-domain}. Differently, some recent methods adopted BERT to encode event sentences~\cite{liu-etal-2020-knowledge,zhou2020temporal}. For example, Liu et al.~\cite{liu-etal-2020-knowledge} proposed knowledge enhanced event causal relation extraction with masking generalizations. The model consists of the knowledge-aware reasoner, masking reasoner, and attentive sentinel trading off between them. The first module uses BERT to model sentences, where events are replaced by definitions from external knowledge, learning expressive event embeddings. The second one also applied BERT on sentences, but event mentions are replaced by [MASK] symbol, mining event-agnostic and context-specific patterns.

Recently, some BERT-based studies tackled data lacking in causal relation extraction via introducing external knowledge \cite{zuo-etal-2020-knowdis,zuo-etal-2021-learnda,zuo-etal-2021-improving,cao-etal-2021-knowledge}. Zuo et al.~\cite{zuo-etal-2020-knowdis} proposed a knowledge enhanced distant data augmentation framework. They extracted causal event pairs based on lexical knowledge and used the results to label sentences distantly, which were further refined with causal commonsense knowledge. Then, they employed relabeling and annealing strategies to make use of distantly labeled sentences to train the causal relation extraction model based on the BERT encoder and FCN-based classifier. Zuo et al.~\cite{zuo-etal-2021-learnda} further proposed a knowledge-guided and learnable data augmentation framework. They regarded causal relation extraction and sentence generation as dual tasks and modeled the mutual relations via dual learning. The generation was initialized with causal event pairs from external knowledge, ensuring the causality of generated sentences. Both causal relation extraction and sentence generation therein were based on BERT. Differently, Zuo et al.~\cite{zuo-etal-2021-improving} learned context-specific causal patterns from external causal statements. Then, they adopted a contrastive transfer strategy to incorporate the learned patterns into the target causal relation extraction model based on BERT. Cao et al.~\cite{cao-etal-2021-knowledge} concatenated contextualized embeddings from the BERT of event sentences, GCN embeddings of one-hop neighbors of events from external knowledge, and densely connected GCN embeddings of the shortest multi-hop paths between events from external knowledge. They then passed the concatenated embeddings to the FCN-based classifier.

The above BERT-based causal relation extraction methods are limited to the intra-sentence setting. For document-level setting, Phu and Nguyen~\cite{tran-phu-nguyen-2021-graph} proposed a graph-based model. They applied BERT to encode words in the document, which were used to generate an interaction graph for the document considering the discourse, syntax, and semantic information. This graph was then consumed by GCN to learn document context-augmented embeddings for causality extraction based on FCN and softmax.

\textbf{In general,} existing event relation extraction methods (see Table~\ref{table:ere_comparison}) cannot fully satisfy the requirements of EKG construction. For example, they usually only focus on verbs as events and do not consider nouns. Actually, event triggers can be verbs or nouns. Another limitation is that they ignore arguments. Future researches on event relation extraction should pay attention to these fundamental problems.

\subsubsection{Event Coreference Resolution}
There are usually many texts describing the same events. It is necessary to group the events referring to the same real-world event into the same cluster after event extraction. This task is called event coreference resolution. Formally, 
\begin{myDef}
\textbf{Event coreference resolution:} Given a set of texts $\{T_0, T_1, ..., T_{l}\}$, each text $T_i$ ($0\leq i \leq l$) contains some events $\{e_{i}^{0}, e_{i}^{1}, ..., e_{i}^{\alpha}\}$, it divides these events into clusters $\{C_{0}, C_{1}, ..., C_{\beta}\}$, where each cluster consists of events that refer to the same real-world event.
	\label{def:ECR}
\vspace{-1mm}
\end{myDef}
For example, 
\begin{myExam}
    Input: $T_0$: Barack Obama previously \underline{served as} a U.S. senator from 3 January 2005. Then, he was \underline{selected} \underline{as} the President of America on 20 January 2009.\\
    $T_1$: Before \underline{winning} the presidential election, Obama \underline{represented} Illinois in the U.S. Senate from 2005 to 2008.
	Output: Two event clusters: $C_0$ = \{served as, represented\} and $C_1$ = \{selected as, winning\}.
	\label{exam:ECR}
\vspace{-1mm}
\end{myExam}

\begin{figure}[!htb]
	\centering
	\vspace{-3mm}
	\includegraphics[width=3.3in]{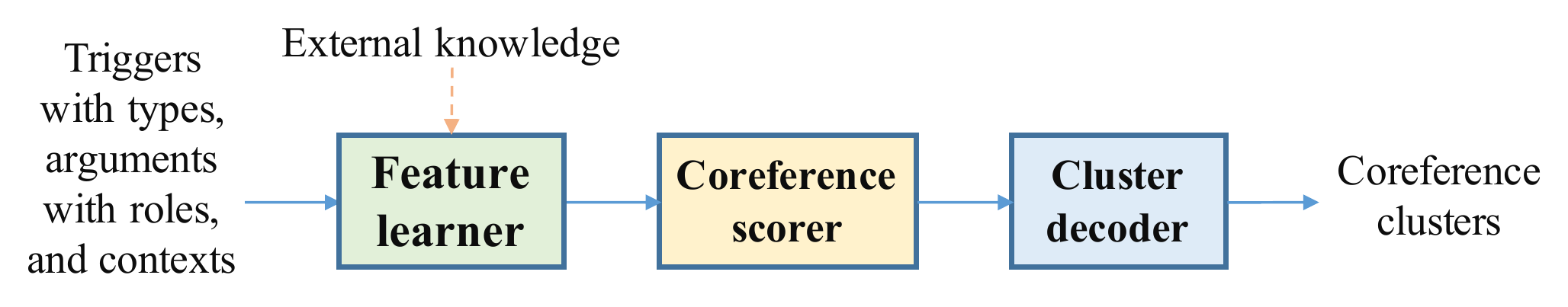}
	\vspace{-4mm}
	\caption{Existing formalization of event coreference resolution.}
	\label{fig:ECR_paradigm}
	\vspace{-1mm}
\end{figure}

Event coreference resolution is divided into within- and cross-document settings according to whether events are from the same document or different ones. The latter is more intricate, since it is difficult to deal with event contexts from different documents. For example, semantically similar event contexts from different documents may describe different events. As illustrated in Fig.~\ref{fig:ECR_paradigm}, existing methods pass the results from event extraction and their contexts to the feature learner and coreference scorer to get the coreference results between events. Then, a cluster decoder is applied to merge the local results to get the global ones, where some rules or clustering algorithms are adopted. Some methods additionally introduce external knowledge to improve feature learning. Specifically, existing methods can be divided into unsupervised, semi-supervised, and supervised ones.

\begin{table*}[htb]
	\setlength{\tabcolsep}{1em}
	\centering
	\caption{Comparison of methods for event coreference resolution.}
	\vspace{-1mm}
	\vspace{-2mm}
	\makebox[\columnwidth][c]{
		\begin{tabular}{c|c|c|c}
			\toprule[1pt]
			Method category & Description & Strengths & Weaknesses\\
			\midrule[0.5pt]
			\makecell[c]{Unsupervised} &\makecell[c]{Construct feature template-based event \\\!\!representations and perform event matching\!\!} &Intuitive and efficient &Limited scalability  \\
			\midrule[0.5pt]
			\makecell[c]{Semi-supervised} &\makecell[c]{Make use of unlabeled data or external \\resources to enhance the coreference scorer} &\makecell[c]{Low labor costs} &\makecell[c]{Not so effective and external\\ \!\!resources may introduce noises\!\!} \\
			\midrule[0.5pt]
			\makecell[c]{Supervised \\(event-pair models)} &\makecell[c]{Process event pairs and adopt a binary\\ classifier as the coreference scorer} &\makecell[c]{Effective and handle the\\\!\!cross-document setting well\!\!} &\makecell[c]{High computational\\ complexity}\\
			\midrule[0.5pt]
			\makecell[c]{Supervised \\\!\!(event-ranking models)\!\!} &\makecell[c]{Process the antecedent events simultaneously\\ and find the first coreferential antecedent} &\makecell[c]{Use more contexts} &\makecell[c]{Less suitable for the \\cross-document setting}\\
			\bottomrule[1pt]
		\end{tabular}
	}
	\vspace{-3mm}
	\label{table:ECR_comparison}
\end{table*}

\textbf{Unsupervised methods} construct feature template-based event representations and adopt pattern matching or unsupervised probabilistic models to identify coreference relations \cite{whittemore1991event,humphreys1997event,chen2009graph,bejan-ECB:2010,cybulska2014using,chen2015chinese,ribeiro2017unsupervised,zhukova2021xcoref}.

Early researches adopted rule-based approaches to deal with event coreference resolution \cite{whittemore1991event,humphreys1997event}. They applied the consistency of event triggers and arguments to determine whether two events are coreferential. Subsequently, some studies \cite{chen2009graph,bejan-ECB:2010,zhukova2021xcoref} used lexical features of event triggers, argument-related features, semantic features, and other handcrafted ones to construct event representations. Then, maximum entropy, non-parametric Bayesian estimation, cosine similarity, and other methods were adopted to determine the coreference of event pairs. Moreover, Chen and Ng~\cite{chen2015chinese} proposed an unsupervised probabilistic model for event coreference resolution and further introduced anaphoricity determination. The lexical and semantic features of triggers and arguments were also applied, and the expectation-maximization algorithm was used to estimate the model's parameters. These unsupervised methods can usually handle both within- and cross-document settings.

\textbf{Semi-supervised methods} pay attention to the scarcity of existing annotated corpus~\cite{raghavan2012exploring,sachan2015active,peng2016event,choubey-huang-2021-automatic}. They use a small amount of labeled data and a large amount of unlabeled data to conduct event coreference resolution. 

For example, Sachan et al.~\cite{sachan2015active} proposed an active learning-based event coreference resolution method. Some heuristic sample selection strategies, such as the maximum uncertainty, maximum expected judgment error, and exploration and exploitation, were used to choose event pairs for manual annotating. Peng et al.~\cite{peng2016event} conducted trigger detection and event coreference resolution in a unified framework. For the former, the similarity between candidate trigger and event type was adopted, where event type embedding is the embedding average of its event examples. For the latter, the similarity between event embeddings was applied. Event embedding was obtained via concatenating the embeddings of its elements trained from external texts. Their similarity thresholds were tuned by the given few labeled samples. These semi-supervised methods can usually be applied to both within- and cross-document settings. 

\textbf{Supervised methods.} With the construction of datasets such as MUC~\cite{MUC}, ACE~\cite{walker2006ace}, and ECB/ECB+~\cite{bejan2008linguistic,cybulska2014using}, and the development of the TAC KBP Event Nugget Detection evaluation task~\cite{song2015light}, researchers have developed many supervised methods for event coreference resolution. They are divided into event-pair and event-ranking models depending on the sample form of the coreference scorer.

\textit{Event-pair models} are common and influential. They process event pairs and adopt binary classifier as the coreference scorer to assign each event pair a probability of being coreferential \cite{lee:2012,CV2015a-bag:2015,krause2016event,Choubey-Iterate:2017,kenyon2018resolving,barhom:2019,zeng2020event}.

For example, Krause et al.~\cite{krause2016event} adopted CNN as the feature learner to process event sentences, whose outputs were concatenated with the embeddings of event triggers and their left and right neighbors to get event embeddings. Two event embeddings were concatenated and augmented with their features, then passed to an FCN, followed by a logistic regression classifier as the coreference scorer. This method only focuses on within-document event pairs. To handle both within- and cross-document settings, Choubey and Huang~\cite{Choubey-Iterate:2017} separately trained two neural network-based feature learners and FCN-based coreference scorers. Then, they alternated within- and cross-document cluster merging to model the second-order inter-dependencies across events. To further capture the semantic interactions between event contexts, Zeng et al.~\cite{zeng2020event} proposed an interaction-based within- and cross-document coreference model. Specifically, two sentences of the event pair were concatenated and fed to the feature learner based on BERT. Meanwhile, the internal structure of the events was injected via semantic role label embeddings. FCN and softmax were used as the coreference scorer. Lee et al.~\cite{lee:2012} and Barhom et al.~\cite{barhom:2019} further introduced entity coreference resolution to enhance the performance by interacting the two tasks. Lexical resource or pretrained word embeddings were used therein.

\textit{Event-ranking models} process all events mentioned before the given event, i.e., the antecedents, simultaneously. They are trained to rank the first coreferential antecedent of each given event first \cite{araki2015joint,lu2017learning,lu2017joint,lu2020end,lu2021span,Tran2021Exploiting,lu2021constrained}.

For example, Lu and Ng~\cite{lu2017learning} selected the coreferential antecedent for each event in a document collectively. They defined the antecedent vector, where the $i$-th element is the coreferential antecedent index of the $i$-th event in the document. A log-linear model was employed to assign the highest score to this antecedent vector. Since coreference may be long-distance, document-level information is helpful. Thus, Tran et al.~\cite{Tran2021Exploiting} constructed a structure graph of the events, entities, and words for each document, and applied GCN on it. Event embeddings from GCN were fed to the FCN-based coreference scorer. To make use of the cross-task interactions for better performance, many methods further incorporated trigger detection, entity coreference resolution, anaphoricity determination, realis detection, or argument extraction into a joint learning framework \cite{araki2015joint,lu2017joint,lu2020end,lu2021span,lu2021constrained}.

Notably, since event-ranking models usually need more contexts, they are more suitable for the within-document setting and may introduce noises for the cross-document one.

\textbf{Generally speaking,} existing researches on event coreference resolution (see Table~\ref{table:ECR_comparison}) still have some deficiencies. For example, most methods specify that all events have fixed arguments. However, arguments differ event by event. In addition, some methods only consider the within-document setting and cannot handle the cross-document one, while both are important for better EKG construction. Thus, future researches should develop practical methods.

\begin{table*}[htb]
	\setlength{\tabcolsep}{1em}
	\centering
	\caption{Comparison of methods for event argument completion.}
	\vspace{-1mm}
	\vspace{-2mm}
	\makebox[\columnwidth][c]{
		\begin{tabular}{c|c|c|c}
			\toprule[1pt]
			Method category &Description &Strengths &Weaknesses\\
			\midrule[0.5pt]
			\makecell[c]{Learn from \\thematic fit} &\makecell[c]{Combine known elements and further the target one\\ based on neural networks to predict the missing one} &Simple &Not so effective\\
			\midrule[0.5pt]
			\makecell[c]{Based on \\graph sequence} &\makecell[c]{Design the score learner upon the \\time-specific embeddings of event elements} &Good at interpretability &Need time to be known \\
			\midrule[0.5pt]
			\makecell[c]{From n-ary facts} &\makecell[c]{Evaluate the composability of event elements} &Good at interpretability &High model complexity \\
			\bottomrule[1pt]
		\end{tabular}
	}
	\vspace{-3mm}
	\label{table:EAC_comparison}
\end{table*}
\subsubsection{Event Argument Completion}
Since information in original texts is incomplete, and there are some missing in event extraction, the extracted events usually miss some elements. Event argument completion thus aims to complete existing events, generally formalized as filling in a missing argument or argument role. Formally,
\begin{myDef}
\textbf{Event argument completion:} Given the EKG $G$, it is to fill in event information for each incomplete event $e$, including filling in the argument $arg^*$ for a target argument role $tgt_{rl}$, $arg^* = \arg \max_{\forall arg} \Pr(arg | e, tgt_{rl}, G)$ and filling in the argument role $rl^*$ for a target argument $tgt_{arg}$, $rl^* = \arg \max_{\forall rl} \Pr(rl | e, tgt_{arg}, G)$.
	\label{def:EAC}
\vspace{-1mm}
\end{myDef}
For example,
\begin{myExam}
	Input: \{event type: Start-position, person: Barack Obama, position: U.S. senator\}, target argument role ``time'', and other existing events.\\
	Output: 3 January 2005.
	\label{exam:EAC}
\vspace{-1mm}
\end{myExam} 

\begin{figure}[!htb]
	\centering
	\vspace{-2mm}
	\includegraphics[width=2.9in]{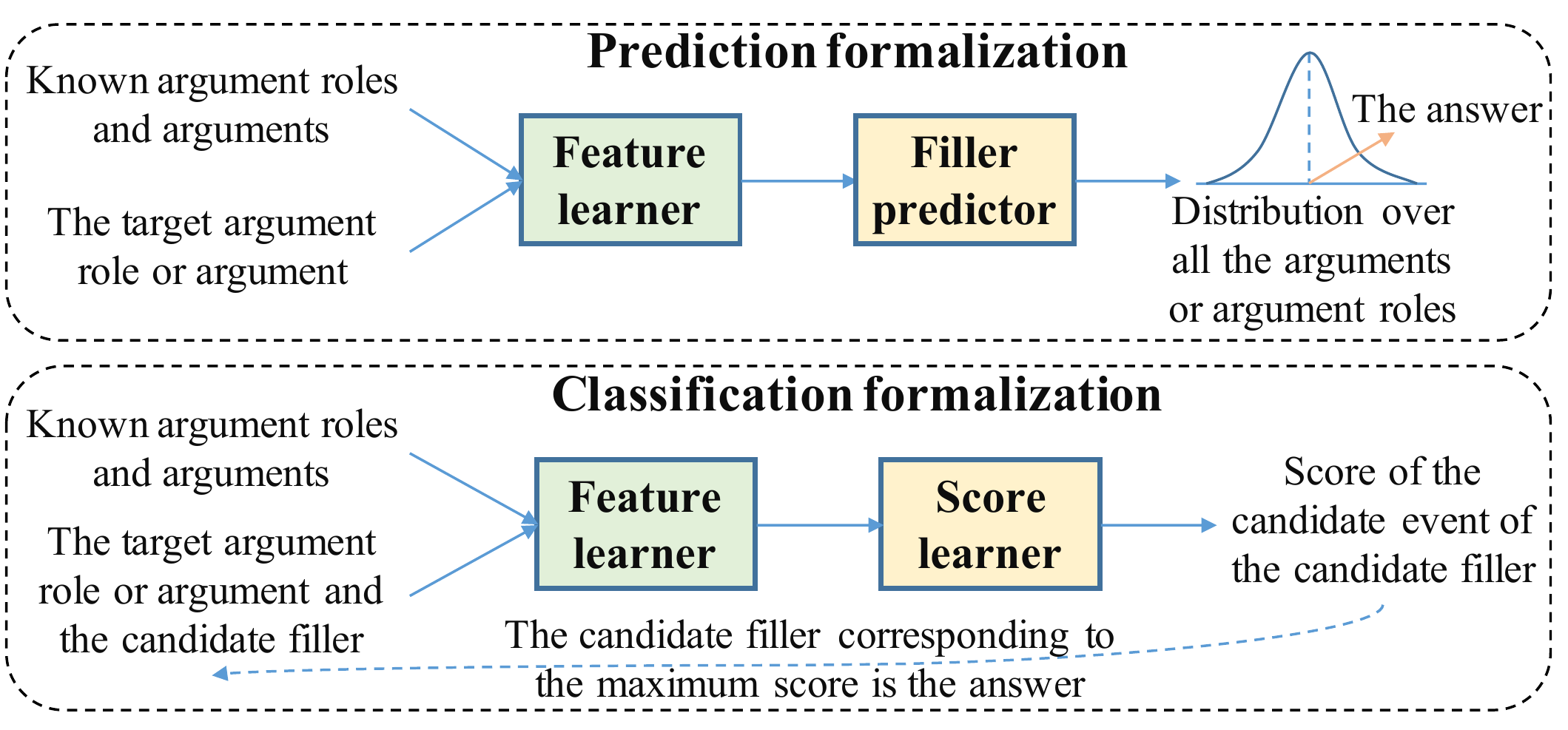}  
	\vspace{-3mm}
	\caption{Existing formalizations of event argument completion.}
	\label{fig:EAC_formalization}
	\vspace{-2mm}
\end{figure}

Existing methods further formalize this task as a prediction or classification task. As shown in Fig.~\ref{fig:EAC_formalization}, the former passes known elements to the feature learner and filler predictor to get the predicted element. It is compared with all the arguments or argument roles to get the overall distribution with the peak as the answer. The latter also takes the candidate filler as input and learns the score of the candidate event. The candidate filler of the maximum score is the answer.

\textbf{Some early researches learned from thematic fit for sentence comprehension,} which determines the goodness of fit between the entities and the agent and patient roles of the verb \cite{thematic_fit}. For example, Tilk et al.~\cite{EAC16} combined the embeddings of argument role-argument pairs based on neural networks. They added these combined embeddings of known pairs to further combine with the target argument role and then predicted the missing argument. Hong et al.~\cite{EAC18} replaced the sum of the pair embeddings with their weighted sum and additionally predicted the missing argument role for the target argument. 

\textbf{EKG with time can be reorganized as a sequence of graphs by time. Event argument completion is based on the score learner upon the time-specific embeddings of event elements in the graph sequence.} Garc{\'\i}a-Dur{\'a}n et al.~\cite{garcia2018learning} and Leblay and Chekol~\cite{leblay2018deriving} integrated the time into event type embeddings by concatenating their embeddings. Then, the score learner was designed on these time-aware embeddings of event types and the embeddings of other arguments as their combination like TransE~\cite{bordes2013translating}. Dasgupta et al.~\cite{dasgupta2018hyte} assigned each timestamp with a hyperplane. Events valid at a certain time were projected onto the corresponding hyperplane, where the translational scores of arguments via event type~\cite{bordes2013translating} were used for completion. Xu et al.~\cite{xu2020time} further maintained the smoothness between adjacent hyperplanes by minimizing their Euclidean distances. Differently, Lacroix et al.~\cite{lacroix2020tensor} represented an EKG as a tensor of event types, time, and other arguments. They then performed tensor decomposition and used the reconstructed tensor to conduct completion. Besides events with exact occurring time, TeRo~\cite{xu2020tero} additionally handled events with beginning and end time, and accordingly represented each event type as a pair of dual complex embeddings. They mapped other arguments to these two event type embeddings and combined the translation-based scores~\cite{bordes2013translating}.

\textbf{Events are typical n-ary facts. Completion methods for n-ary facts can also be used to event argument completion.} Early studies applied translation-based methods. Wen et al.~\cite{m-TransH} defined the score learner for each candidate event as the weighted sum of the projection results from its arguments to the hyperplane of its event type, where the weights are the real numbers projected from its argument roles. Zhang et al.~\cite{RAE} additionally introduced the likelihood that two arguments co-participate in a common event via FCN. Liu et al.~\cite{RAM} further considered the relatedness among argument roles and the compatibility between each argument role and all the involved arguments. Subsequently, tensor-based methods were adopted. Liu et al.~\cite{GETD} represented events as a tensor and reconstructed it to make completion. Di et al.~\cite{S2S} further solved data sparsity via partially sharing embeddings and over-parameterization by sparsifying the tensor. Recently, neural network-based methods have sprung up. They used CNN \cite{NaLP,HypE,HINGE,tNaLP_plus}, FCN \cite{NaLP,NeuInfer,tNaLP_plus}, GNN \cite{STARE,CoRelatE}, and Transformer \cite{STARE,GRAN} to learn features and obtain scores of candidate events or fillers. For example, Guan et al.~\cite{NaLP} adopted CNN to get the embeddings of argument role-argument pairs. The relatedness of these pairs via FCN was used to estimate the scores of candidate events. Galkin et al.~\cite{STARE} organized events into a graph and used GCN as the feature learner to get the embeddings of event elements. The learned embeddings of known elements were passed to a Transformer-based filler predictor to get answer distribution.

\begin{table*}[htb]
	\setlength{\tabcolsep}{1em}
	\centering
	\caption{Summarization of EKG-related representative graphs/systems.}
	\vspace{-1mm}
	\vspace{-2mm}
	\makebox[\columnwidth][c]{
		\begin{tabular}{c|c|c|c}
			\toprule[1pt]
			Graph/system &Domain &Description &Scale\\
			\midrule[0.5pt]
			\makecell[c]{WikiNews, FIFA WorldCup, \\Cars, and Airbus Corpus~\cite{ECKGs16}} &\makecell[c]{\!\!General news, football,\!\! \\automotive industry, \\and airbus A380} &\makecell[c]{Focus on event temporal and causal \\relations, and events have actions,\\ participants, time, and locations} &\makecell[c]{Over 624 thousand, 9.3 \\million, 25 million, and 2.5 \\thousand events, respectively}\\
			\midrule[0.5pt]
			\makecell[c]{\!\!Chinese travel domain event\!\! \\evolutionary graph~\cite{EEG17}} &Tourism &\makecell[c]{Temporal relations between\\ events (verb phrases)} &-\\
			\midrule[0.5pt]
			\makecell[c]{Chinese financial domain \\event logic graph~\cite{ELG19}} &Finance &\makecell[c]{Temporal, causal, conditional and \\hypernym-hyponym relations \\between events ($(s, p, o)$ tuples)} &\makecell[c]{Over 1.5 million \\events and their \\1.8 million relations}\\
			\midrule[0.5pt]
			\makecell[c]{Event-centric \\Hainan tourism KG~\cite{Hainan20}} &Tourism &\makecell[c]{Temporal and spatial \\dynamics of tourists’ trips} &\makecell[c]{Over 7 thousand journeys, \\about 87 thousand events, \\about 141 thousand entities, \\and near 228 thousand edges \\of event arguments and about 80 \\\!\!thousand event temporal relations\!\!}\\
			\midrule[0.5pt]
			EventPlus~\cite{EventPlus21} &Multiple domains &\makecell[c]{System: Extract event triggers, argu-\\ments, duration, and temporal relations} &-\\
			\midrule[0.5pt]
			\makecell[c]{EventKG~\cite{EKG18}, EventKG+\\Click~\cite{EventKG_Click20}, and OEKG~\cite{OEKG21}} &General domain &\makecell[c]{Focus on subevent, previous event, and \\next event relations, and events have \\topics, participants, time, and locations} &\makecell[c]{Over 690 thousand events, -, \\and over 436 million triples}\\
			\midrule[0.5pt]
			CogCompTime~\cite{CogCompTime18} &General domain &\makecell[c]{System: Extract time expression, event \\triggers, and trigger temporal relations} &-\\
			\midrule[0.5pt]
			ASER~\cite{ASER20} &General domain &\makecell[c]{Temporal, contingency, comparison, \\expansion, and co-occurrence between \\eventualities (dependency graphs)} &\makecell[c]{Over 194 million eventualities \\and 64 million relations}\\
			\bottomrule[1pt]
		\end{tabular}
	}
	\vspace{-3mm}
	\label{table:graph_system_summarization}
\end{table*}

\textbf{On the whole}, as summarized in Table~\ref{table:EAC_comparison}, existing studies focus on event itself, while ignoring event relations. Introducing event relations may be helpful. It is an interesting direction in the future. Moreover, researchers simplify event argument completion as inferring a missing argument or argument role. However, the argument role and its argument are usually both missing. Thus, future studies should pay attention to more realistic formalizations and methods.

\subsection{EKG-related Representative Graphs/Systems}
With the development of event acquisition, there are some EKG-related representative graphs/systems targeting a specific or general domain.

\textbf{Domain-specific graphs/systems.} In 2016, Rospocher et al.~\cite{ECKGs16} constructed four event-centric KGs, i.e., WikiNews in English, FIFA WorldCup in English, Cars in English, and Airbus Corpus in English, Dutch, and Spanish, from different news. They have over 624 thousand, 9.3 million, 25 million, and 2.5 thousand events, respectively. Some specific event-event and event-entity relations are considered as introduced in Section~\ref{subsec:EKG_history}. In 2017, Li et al.~\cite{EEG17} constructed a Chinese travel domain event evolutionary graph from a large-scale unstructured Web corpus. Its nodes are events (simplified as verb phrases), and edges are sequential relations between events with transition probability. In 2019, Ding et al.~\cite{ELG19} constructed a Chinese financial domain event logic graph from plenty of news. It focuses on event causal relations and has over 1.5 million event nodes (i.e., $(s, p, o)$ tuples) and their 1.8 million directed edges. In 2020, Wu et al.~\cite{Hainan20} constructed an event-centric tourism KG based on touristic data in Hainan to model the temporal and spatial dynamics of tourists' trips. Its nodes are over 7 thousand journeys, about 87 thousand events, and about 141 thousand entities, while its near 228 thousand edges are event arguments and about 80 thousand event temporal relations. Each event contains three components, activity, time, and place, and is connected to its journey via the relation ``contain''. In 2021, Ma et al.~\cite{EventPlus21} presented the event pipeline system EventPlus\footnote{See https://kairos-event.isi.edu/}, with comprehensive event understanding capabilities to extract event triggers, arguments, duration, and temporal relations. It was designed with multi-domain support by multi-domain training. However, in its temporal relation graph, nodes are event triggers, and edges are their temporal relations. The event arguments are separate and not presented in the graph.

\textbf{General-domain graphs/systems.} In 2018, Gottschalk and Demidova~\cite{EKG18} constructed a multilingual event-centric temporal KG EventKG from structured and semi-structured data, and some event-event, event-entity, and entity-entity relations were considered (see Section~\ref{subsec:EKG_history}). It has over 690 thousand events. It was further extended to EventKG+Click~\cite{EventKG_Click20} by introducing user interactions with events, entities, and their relations, derived from the Wikipedia clickstream. Besides, based on EventKG, Gottschalk et al.~\cite{OEKG21} built OEKG (Open EKG) with over 436 million triples by further integrating event-related data sets from multiple application domains, such as question-answering, entity recommendation, and named entity recognition. Also in 2018, Ning et al.~\cite{CogCompTime18} proposed the temporal understanding system CogCompTime to extract time expression, event triggers, and temporal relations between event triggers. Argument information was not considered therein. In 2020, Zhang et al.~\cite{ASER20} developed ASER (Activities, States, Events, and their Relations), an English eventuality KG extracted from reviews, news, forums, social media, movie subtitles, and e-books. In ASER, each node is an eventuality, which is a dependency graph, and each edge is a relation between eventualities. In the dependency graph, nodes are the words in the sentence, and edges are their dependency relations. Five categories of eventuality relations were considered, i.e., temporal relations, contingency, comparison, expansion, and co-occurrence. The full version of ASER has over 194 million eventualities and 64 million relations.

\textbf{Thus,} there are some EKG-related representative graphs or systems, summarized in Table~\ref{table:graph_system_summarization}. However, they all consider specific and limited event-event relations or argument roles. Actually, there are various event-event relations in real-world scenarios. Moreover, different events usually do not share argument roles. There is a need to develop practical EKGs to facilitate downstream applications in the future.

\section{What is EKG: Application View}
\label{sec:application_view}
EKG has substantial application value, with the introduction of events and event-event relations. This section introduces its basic and deep applications.

\subsection{Basic Applications}
The main basic applications are predictions on EKG, which predict future events based on the current EKG. There are two ways to deal with these predictions. The first one generalizes event instances to script events and predicts the subsequent script events at the script level, called script event prediction \cite{chambers-jurafsky-2008-unsupervised}. The predicted script events can then be instantiated to real-world events. The second one predicts future events at the instance level directly. Specifically, existing methods simplify EKG to temporal KG, formalized as a sequence of KGs of $(s,p,o)$ with timestamps. Then, future prediction is to predict events for future timestamps, called temporal KG prediction \cite{trivedi2017know,trivedi2018dyrep,jin2019recurrent,jin2020Renet}.

\subsubsection{Script Event Prediction}
\label{subsubsec:script_event_prediction}
\begin{figure}[!htb]
	\centering
	\vspace{-6mm}
	\includegraphics[width=2.5in]{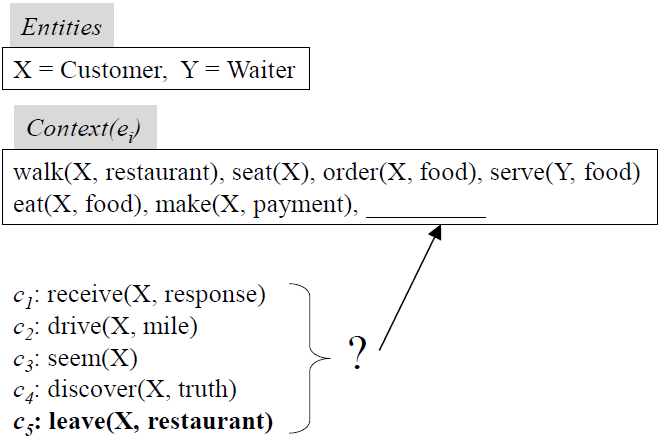}
	\vspace{-3mm}
	\caption{A typical example of script event prediction~\cite{wang-etal-2017-integrating}.}
	\vspace{-3mm}
	\label{fig:SEP_setting}
\end{figure}

Script event prediction is to predict the subsequent script events given the historical scripts. In the following, script event is denoted as event for convenience. Formally, 
\begin{myDef}
    \textbf{Script event prediction:} Given the historical events $\mathcal{H}=\{e_0, e_1, ..., e_{x}\}$, script structure $\mathcal{S}=\{r_0(e_{h_0}, e_{t_0}), r_1(e_{h_1}, e_{t_1}), ..., r_y(e_{h_y}, e_{t_y})\}$, and candidate events $\mathcal{C}=\{e_{c_0}, e_{c_1}, ..., e_{c_z}\}$, it predicts the most possible subsequent event $e^* = \arg \max_{e\in \mathcal{C}} \Pr(e | \mathcal{H}, \mathcal{S}) $.
    \label{def:SEP}
\vspace{-1mm}
\end{myDef}
A typical example is presented in Fig.~\ref{fig:SEP_setting}, where $Context(e_i)$ is the script structure for the subsequent event $e_i$ to be predicted. The relations are all temporal one and omitted. 

\begin{table*}[htb]
	\setlength{\tabcolsep}{1em}
	\centering
	\caption{Comparison of methods for script event prediction.}
	\vspace{-1mm}
	\vspace{-2mm}
	\makebox[\columnwidth][c]{
		\begin{tabular}{c|c|c|c}
			\toprule[1pt]
			\!\!\!Method category\!\!\! &Description &Strengths &Weaknesses\\
			\midrule[0.5pt]
			\!\!\!Chain-modeling\!\!\! & 
			\makecell[c]{\!\!\!Model event pairs or chain\!\!\!} & 
			\makecell[c]{Good at modeling behavior trends} &
			\makecell[c]{\!\!\!Miss information from script events of other entities\!\!\!} \\
			\midrule[0.5pt]
			\!\!\!Graph-modeling\!\!\! & 
			\makecell[c]{Apply GNN models} & 
			\makecell[c]{\!\!\!Catch more comprehensive contexts\!\!\!} &
			\makecell[c]{\!Difficult to capture behavior trend of each entity\!} \\
		    \bottomrule[1pt]
		\end{tabular}
	}
	\vspace{-3mm}
	\label{table:sep_comparison}
\end{table*}

Script event prediction is formalized as script coherence evaluation, and the event corresponding to the max coherence score is chosen as the answer. Chambers and Jurafsky~\cite{chambers-jurafsky-2008-unsupervised} represented events as $(subject, predicate)$ or $(predicate, object)$ pairs. Given an event, they obtained its coherence score by aggregating its coherence scores with all the events from the script based on PMI. The narrative cloze test was adopted for evaluation. In this test, one event in the script is masked, and the model is asked to predict it. Following studies improve this model in three aspects:

\textbf{Event representation.} The pair representation in \cite{chambers-jurafsky-2008-unsupervised} loses the co-occurrence information of subjects and objects. Thus, Balasubramanian et al.~\cite{balasubramanian-etal-2013-generating} applied $(subject,$ $predicate, object)$ triples. Pichotta and Mooney~\cite{pichotta-mooney-2014-statistical} and Granroth-Wilding and Clark~\cite{ERL16_2} further added in indirect object, i.e., $(subject, predicate, object, indirect\ object)$. To solve sparsity in these symbolic representations, the following studies learned distributed event representations, i.e., event embeddings, by composing their components. Relatively early studies applied additive composition methods \cite{ERL16,wang-etal-2017-integrating,bl_emnlp21}. They added linearly transformed predicate and argument embeddings. Bai et al.~\cite{bl_emnlp21} further added the embeddings of event sentences. Then, Weber et al.~\cite{ERL18} and Ding et al.~\cite{ERL19} proposed tensor-based models to capture multiplicative interactions between event elements. To introduce external knowledge, Ding et al.~\cite{ERL19} further learned the intent and sentiment of the event. Recently, other methods were adopted, such as probabilistic models~\cite{ERL_skipgram18} and FCN \cite{ERL16_2,lee-goldwasser-2019-multi}. Lee and Goldwasser~\cite{lee-goldwasser-2019-multi} further introduced event discourse relations via their composition to learn relation-aware event embeddings. Besides getting composite event embeddings, Li et al.~\cite{ijcai2018-584} organized the events into a graph and used GNN to update embeddings.

\textbf{Script modeling.} Early studies modeled event pairs in the script, which ignore event order or only consider limited event order of event pair. Different from PMI used in \cite{chambers-jurafsky-2008-unsupervised}, they obtained the coherence score of event pairs via Bigram~\cite{jans-etal-2012-skip,pichotta-mooney-2014-statistical}, FCN~\cite{ERL16_2}, cosine similarity~\cite{ERL18,ERL_skipgram18}, or translation-based composition~\cite{lee-goldwasser-2019-multi}. Other studies modeled the whole event chain in the script via language model~\cite{rudinger-etal-2015-script,peng-etal-2019-knowsemlm}, neural network-based probabilistic model~\cite{ERL16}, or LSTM~\cite{AAAI1612157,wang-etal-2017-integrating,DBLP:conf/aaai/LvQHHH19}. Wang et al.~\cite{wang-etal-2017-integrating} additionally made use of event pair-based models. Different from these single-chain models, Chambers and Jurafsky~\cite{chambers-jurafsky-2009-unsupervised} and Bai et al.~\cite{bl_emnlp21} aggregated the results from multiple event chains. Other than a chain, Li et al.~\cite{ijcai2018-584} and Ding et al.~\cite{ERL19} organized the events into a graph and modeled it via GNN.

\textbf{Evaluation.} The narrative cloze test in \cite{chambers-jurafsky-2008-unsupervised} cannot recognize multiple plausible events, since only the original subsequent event is viewed as the answer. Thus, Modi~\cite{ERL16} proposed the adversarial narrative cloze test. It asks models to distinguish a correct chain from a negative one, a copy of the correct chain with the subsequent event replaced randomly. The Multiple Choice Narrative Cloze (MCNC) test~\cite{ERL16_2} further restricts the subsequent event to a few choices (see the bottom part of Fig.~\ref{fig:SEP_setting}). Based on MCNC, Lee and Goldwasser~\cite{ERL_skipgram18} proposed Multiple Choice Narrative Sequences (MCNS) and Multiple Choice Narrative Explanation (MCNE) to evaluate the ability of inferring long event sequences. MCNS creates candidates for each step. MCNE additionally needs the end event and infers what happened in between. Among these tests, MCNC is widely used.

\textbf{In a word,} as presented in Table~\ref{table:sep_comparison}, there are two ways to represent script. One is entity-centric, organizing the script events into chains according to different participants. The other is the emerging event-centric method, organizing all the script events into a graph. The former is better at modeling the relations between the participant and its script events, while the latter is better at modeling the interactions of script events. Combining their advantages remains to be studied. There are also some other challenges. For example, how to really predict the subsequent script event instead of choosing it from the given candidates? How to combine the information from event instances and script events?

\subsubsection{Temporal KG Prediction}
\label{subsubsec:TKG_prediction}
Temporal KG prediction is to predict events for future timestamps given the historical temporal KG. Formally,
\begin{myDef}
\textbf{Temporal KG prediction:} Given the historical temporal KG in the form of KG sequence with timestamps from $0$ to $t$, i.e., $\mathcal{G}=\{\mathcal{G}_{0}, \mathcal{G}_{1}, ..., \mathcal{G}_{t}\}$, it predicts events $(s,p,o)$ for future timestamp $t+1$, including $o^* = \arg \max_{\forall o} \Pr(o | s,p,t\!+\!1, \mathcal{G})$, $s^* = \arg \max_{\forall s} \Pr(s | p,o,t\!+\!1, \mathcal{G})$, and $p^* = \arg \max_{\forall p} \Pr(p | s,o,t\!+\!1, \mathcal{G})$.
	\label{def:TKGP}
\vspace{-1mm}
\end{myDef}
For example,
\begin{myExam}
	Input: (?, selected as, President of America, 20 January 2009) and historical events before 20 January 2009.\\
	Output: Barack Obama.
	\label{exam:TKGP}
\vspace{-1mm}
\end{myExam} 

\begin{figure}[!htb]
	\centering
	\vspace{-2mm}
	\includegraphics[width=2.8in]{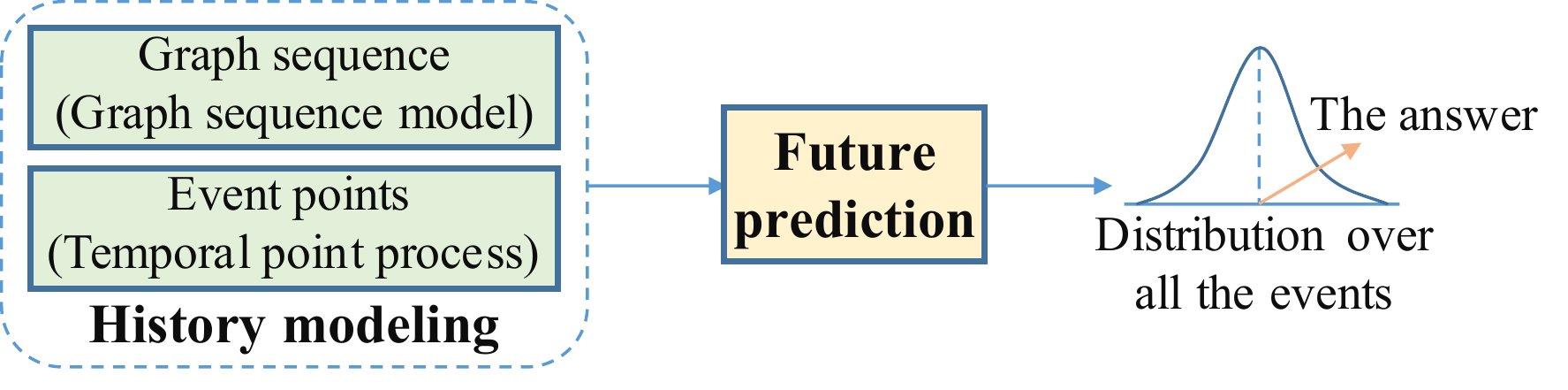}
	\vspace{-3mm}
	\caption{Existing formalization of temporal KG prediction.}
	\label{fig:TKGP_formalization}
	\vspace{-2mm}
\end{figure}

Temporal KG prediction requires models to understand historical events. Thus, as shown in Fig.~\ref{fig:TKGP_formalization}, existing methods formalize it into history modeling and future prediction. They model historical events and their evolution. Based on these, they predict the future event and output a distribution over all the events with the peak as the answer. According to the organization and modeling of historical events, methods can be split into graph sequence-based and temporal point process-based ones. The former organizes historical events as a sequence of graphs, where each graph contains the events that occurred at the corresponding timestamp. The latter considers historical events as event points. 
\begin{table*}[htb]
	\setlength{\tabcolsep}{1em}
	\centering
	\caption{Comparison of methods for Temporal KG prediction.}
	\vspace{-1mm}
	\vspace{-2mm}
	\makebox[\columnwidth][c]{
		\begin{tabular}{c|c|c|c}
			\toprule[1pt]
			\!\!Method category\!\! &Description &Strengths &Weaknesses\\
			\midrule[0.5pt]
			\makecell[c]{Graph \\sequence-based} &\makecell[c]{Predict events based on related\\ historical subgraph sequences} &\makecell[c]{Capture the semantic information of\\ \!concurrent entities and temporal order\!}  & Ignore precise time information\\
			\midrule[0.5pt]
			\makecell[c]{Temporal point \\process-based} &\makecell[c]{Predict events based on the evolut-\\ionary characters of event points} &\makecell[c]{Can model precise time information}  &\makecell[c]{\!High computational complexity and\!\\ not capable for concurrent events}\\
			\bottomrule[1pt]
		\end{tabular}
	}
	\vspace{-3mm}
	\label{table:TKGP_comparison}
\end{table*}

\textbf{Graph sequence-based methods.}
In history modeling, they extract related subgraph sequence. They then learn the evolutional embedding(s) of the graph sequence or subjects, predicates, and objects. GNN and RNN are usually applied for graph and sequence evolution, respectively. Upon the embedding(s), in future prediction, they do argument (i.e., subject/object) prediction (and predicate prediction) via the distribution of all the candidate arguments (or predicates).

Existing methods mainly differ in history extraction. Jin et al.~\cite{jin2019recurrent,jin2020Renet} extracted historical events of the given subject and predicate, or the given subject. Unlike these heuristic methods, Han et al.~\cite{han2020explainable} started from the given subject and iteratively sampled relevant edges of arguments included in the subgraph. Li et al.~\cite{li2021search} and Sun et al.~\cite{sun2021timetraveler} considered history extraction as sequential decision and searched from history via reinforcement learning. Other than the above subgraph sequences, Li et al.~\cite{li2021temporal} used the entire graph sequence from the last several timestamps. These studies model the temporal KG in a discrete-time domain. To encode the continuous dynamics of temporal KG, Ding et al.~\cite{ding2021temporal} extended neural ordinary differential equations~\cite{chen2018neural} to GCN.

\textbf{Temporal point process-based methods.} They introduce conditional intensity function, which is powerful for modeling the evolutionary characters of event points, to do history modeling and future prediction in a unified framework.  

Traditional methods manually specified the conditional intensity function and modeled events without arguments \cite{daley2003introduction,rasmussen2011temporal}. Some recent works~\cite{trivedi2017know,trivedi2018dyrep,han2020graph} extended them to neural methods, using deep neural networks to fit the conditional intensity function. Neural temporal point process-based methods are powerful in considering the semantics of event elements. For example, Trivedi et al.~\cite{trivedi2017know} applied RNN to learn the evolutional embeddings of arguments. They made event argument prediction by estimating the conditional probability of an event based on the evolutional embeddings of the involved arguments and the predicate embedding. That is, the occurrence of an event was modeled as a multivariate point process, whose conditional intensity function was modulated by its score based on the involved embeddings. This method models events occurring continuously, where no events occur at the same timestamp. To further model concurrent events, Han et al.~\cite{han2020graph} aggregated the object embeddings in the concurrent events related to the given subject into a hidden vector. Then, they modeled time as a random variable and deployed the Hawkes process~\cite{hawkes1971spectra} on temporal KG to capture the underlying dynamics, where a continuous-time LSTM was used to estimate the intensity function. 

\textbf{In general,} existing temporal KG prediction methods (summarized in Table~\ref{table:TKGP_comparison}) usually focus on events with three arguments (subject, object, and time) and perform argument or predicate prediction with other elements of future events given. Thus, there are several challenges. For example, how to handle historical events with various arguments? How to practically predict future events on the whole?

\begin{table*}[htb]
	\setlength{\tabcolsep}{1em}
	\centering
	\caption{Summarization of applications on EKG.}
	\vspace{-1mm}
	\vspace{-2mm}
	\makebox[\columnwidth][c]{
		\begin{tabular}{c|c}
			\toprule[1pt]
			Application category &Application description\\
			\midrule[0.5pt]
			Script event prediction &Predict the subsequent script events given the historical scripts (many academic studies)\\
			\midrule[0.5pt]
			Temporal KG prediction &Predict future events given the historical temporal KG (many academic studies)\\
			\midrule[0.5pt]
			Timeline generation &Generate event/biographical timelines using EKG~\cite{TL18,BTG19,BTG20}\\
			\midrule[0.5pt]
			Abductive reasoning &Introduce EKG as additional knowledge into abductive reasoning~\cite{abductive_reasoning21}\\
			\midrule[0.5pt]
			Search &Enable search engines/systems with the searching ability on events based on EKG~\cite{app_search19,app_TCMsearch19}\\
			\midrule[0.5pt]
			Question-answering &Query on events based on EKG~\cite{EventQA20}\\
			\midrule[0.5pt]
			Recommendation &Incorporate additional knowledge from EKG to improve recommendation~\cite{Hainan20}\\
			\midrule[0.5pt]
			Financial quantitative investments &Embeddings from EKG help quantitative trading methods~\cite{ERL20}\\
			\midrule[0.5pt]
			Text generation &Better serve the structured information in EKG in a user-friendly manner~\cite{EventNarrative21}\\
			\bottomrule[1pt]
		\end{tabular}
	}
	\vspace{-3mm}
	\label{table:app_summarization}
\end{table*}

\subsubsection{Other Basic Applications}
There are also direct analyses on EKG, such as timeline generation and abductive reasoning. For example, Gottschalk and Demidova~\cite{TL18} generated cross-lingual event timelines using the multilingual event-centric temporal KG EventKG~\cite{EKG18}. For a query entity or event, they relied on EventKG to provide information concerning the event popularity and relation strength between events and the query\footnote{See http://eventkg-timeline.l3s.uni-hannover.de/}. Gottschalk and Demidova~\cite{BTG19} and Gottschalk and Demidova~\cite{BTG20} handled biographical timeline generation. For a query person, they extracted the most relevant biographical data from EventKG based on event popularity, relation strength, and predicate labels\footnote{See http://eventkg-biographies.l3s.uni-hannover.de/}. Du et al.~\cite{abductive_reasoning21} proposed an event graph enhanced pretrained language model based on variational autoencoder for abductive reasoning, which finds the most reasonable explainable events for the observed events.

\subsection{Deep Applications}
EKG can further facilitate many downstream applications, such as search, question-answering, recommendation, financial quantitative investments, and text generation. For example, Rudnik et al.~\cite{app_search19} developed an event-based search engine\footnote{See https://asrael.eurecom.fr/search-engine-old/home}, able to query both KGs and news articles. Specifically, they mapped events described in news articles to those in Wikidata~\cite{Wikidata}, and attributes from Wikidata were used to annotate the news articles. They then constructed an event-oriented KG and an event-based search engine. Yang et al.~\cite{app_TCMsearch19} implemented a temporal semantic search system for clinical diagnosis and treatment of traditional Chinese medicine. It consists of the offline and online parts. The former is about the construction, storage, and indexing of the temporal KG, and the latter is about the understanding, conversion, and execution of the search sentences. Souza Costa et al.~\cite{EventQA20} addressed answering event-centric questions and constructed the first event-centric question-answering dataset based on the event-centric temporal KG EventKG~\cite{EKG18}. Wu et al.~\cite{Hainan20} proposed a GCN-based method for Point-of-Interest (POI) recommendations, which incorporates tourists' behavior patterns obtained from event-centric tourism KG, to capture the relations between users and POIs effectively. Cheng et al.~\cite{ERL20} presented a KG-based event embedding framework for financial quantitative investments, where the learned embeddings were fed to downstream quantitative trading methods. A mobile mini-app and a Web-based desktop platform were developed based on this framework, obtaining great accumulated portfolio returns. Colas et al.~\cite{EventNarrative21} focused on graph-to-text generation to better serve the structured information in the graph in a user-friendly manner. For each event from EventKG, they augmented the data with additional information from Wikidata and linked the event to a Wikipedia page for text generation.

\textbf{In general,} as summarized in Table~\ref{table:app_summarization}, there are many 
basic and deep applications on EKG. Specifically, studies on script event and temporal KG predictions are richer. Since EKG is a relatively new concept, concrete applications and real-world use cases are in a small number. Thus, exploring more practical applications on EKG is promising.

\section{Future Directions}
\label{sec:future_directions}
There are many researches and achievements on EKG. However, there are still several directions to focus on and investigate further. In this section, we look deep into them.

\subsection{High-performance Event Acquisition}
Recent event acquisition researches are far from meeting the application requirements in effectiveness and efficiency. Especially, the precision of event and event relation extractions is low. Thus, it hinders the construction of high-quality EKGs. Besides, existing models usually do not pay attention to the complexity problem. However, models of high parameter and time complexity go against the fast construction of EKGs from massive data. Thus, highly effective and efficient event acquisition is an essential future direction.

\subsection{Multi-modal Knowledge Processing}
Events are presented in texts, images, audios, and videos in the real world. However, existing EKG researches usually process text, ignoring the information in other modalities. Very few studies look into multi-modal event representation learning \cite{zhang2021MERL} and extraction \cite{li-etal-2020-cross}. Actually, events from different modalities disambiguate and complement mutually. Thus, jointly using multi-modal information is important. Specifically, events from all modalities should be represented in a unified framework, event acquisition should handle multi-modal extraction, and reasoning on EKG should consider the multi-modal information.

\subsection{Interpretable EKG Research}
Existing EKG researches mainly apply deep learning methods to fit the training data. However, they usually lack interpretability, i.e., there are no definite ideas about why and how they work. Actually, knowing the reasons for the final results is useful for adopting them in real applications. It is friendly and convincing to explain why the final results are the given ones. In the future, interpretable EKG research is an important direction.

\subsection{Practical EKG Research}
Currently, the related tasks and methods of EKG are far from real-world scenarios. For the related tasks, some task formalizations are idealized. For example, complete a missing element in an existing event, predict a future script event via choosing it from several candidates, and predict an element for a future event. Researches under more practical formalizations are of great importance for applications. For the methods on EKG, GNN is widely used. However, with the simplified EKG tasks and datasets, these methods employ GNN on EKG similar to that on KG (e.g.,~\cite{ijcai2018-584} and \cite{ding2021temporal}). Actually, EKG consists of events, entities, and their relations. Thus, future methods should pay attention to the atomicity of events with arguments, and the relations between events and between entities.

\section{Conclusions}
\label{sec:conclusions}
EKG is important for many applications, such as search, question-answering, recommendation, financial quantitative investments, and text generation. This paper presents a survey on EKG from different views comprehensively. Specially, we looked deep into the history, ontology, instance, and application views of EKG. Its history, definition, schema induction, acquisition, related representative graphs/systems, and applications are thoroughly studied. Based on the development trends therein, prospective directions are further summarized for future research on EKG.

\ifCLASSOPTIONcompsoc
  \section*{Acknowledgments}
\else
  \section*{Acknowledgment}
\fi

The work was supported in part by the Lenovo-CAS Joint Lab Youth Scientist Project, in part by the Foundation and Frontier Research Key Program of Chongqing Science and Technology Commission under Grant cstc2017jcyjBX0059, in part by the Youth Innovation Promotion Association CAS under Grant 20144310, in part by the National Natural Science Foundation of China under Grants 62002341, U1911401, and 61772501, and in part by the GFKJ Innovation Programs.

\ifCLASSOPTIONcaptionsoff
  \newpage
\fi

\bibliographystyle{IEEEtran}
\bibliography{EKG_survey}

%
%

%

\vspace{-12mm}
\begin{IEEEbiography}
	[{\includegraphics[width=1in,height=1.25in,clip,keepaspectratio]{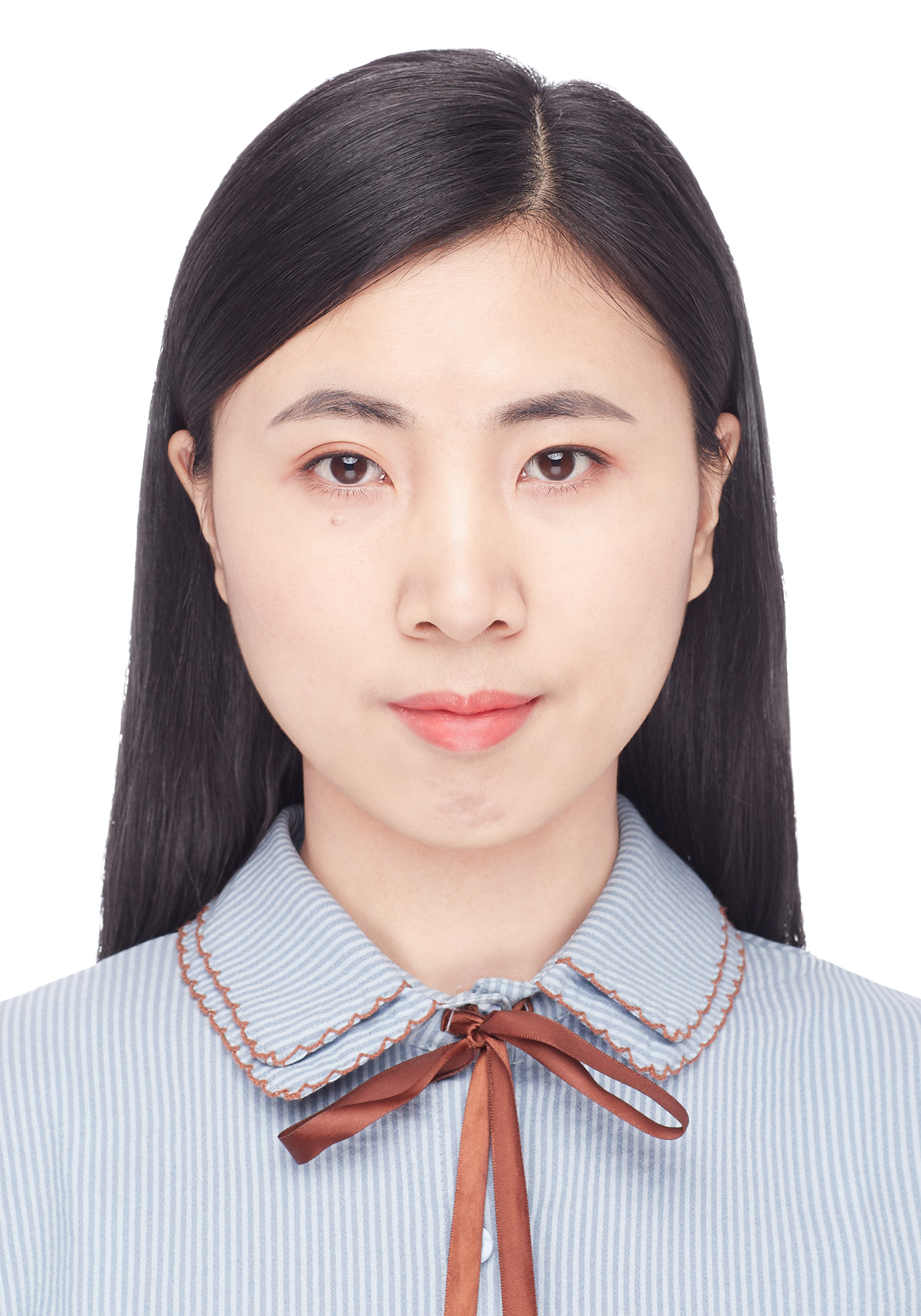}}]{Saiping Guan}
received the PhD degree in Computer Software and Theory in 2019, from the Institute of Computing Technology, Chinese Academy of Sciences, where she is currently an assistant professor. Her current research interests include event knowledge graph, knowledge graph, n-ary relation, etc. She has published papers in prestigious journals and conferences, including TKDE, Knowledge and Information Systems, WWW, ACL, CIKM, etc. She has received the Best Student Paper Award in ICBK (2017).
\end{IEEEbiography}

\vspace{-10mm}
\begin{IEEEbiography}
	[{\includegraphics[width=1in,height=1.25in,clip,keepaspectratio]{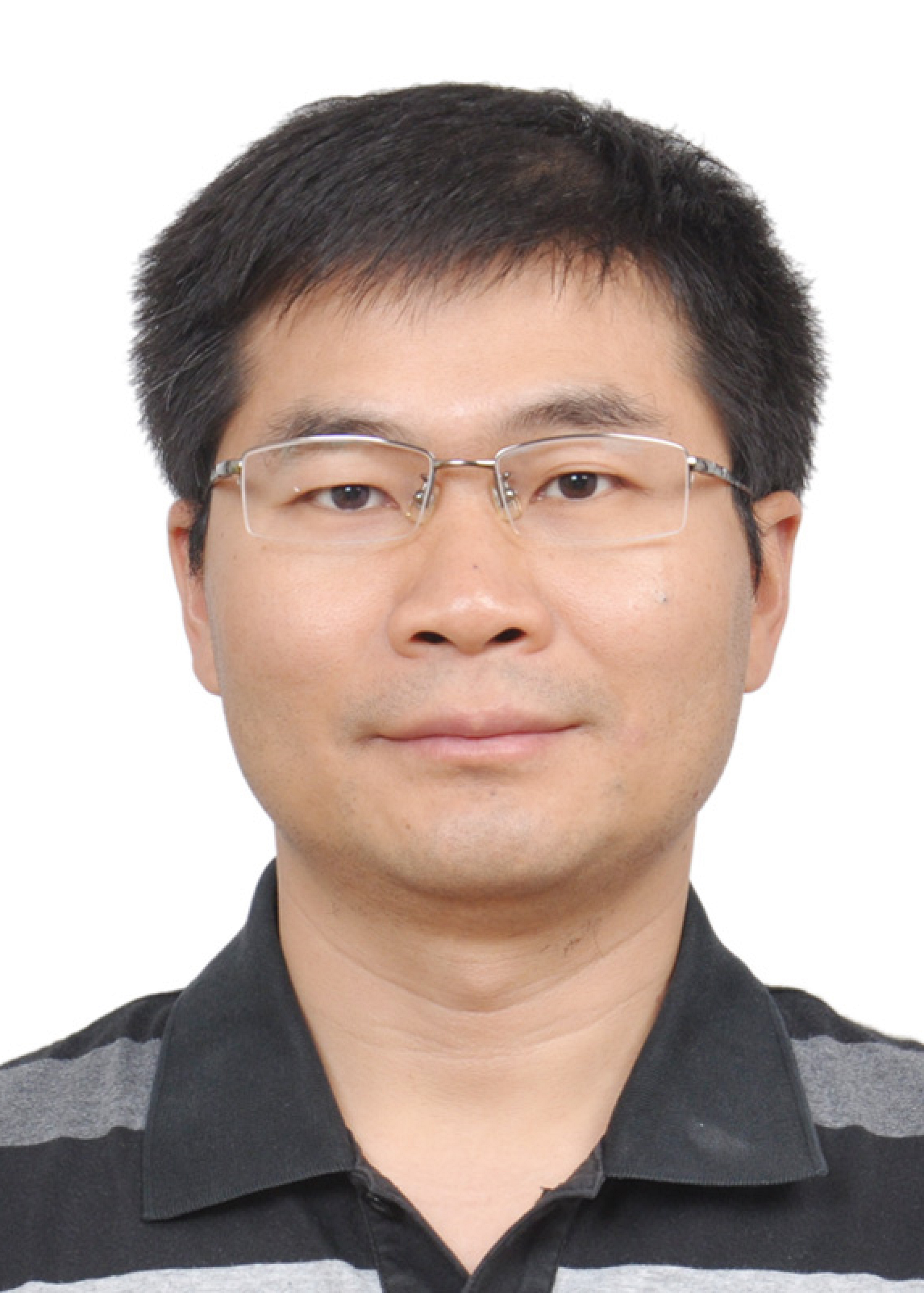}}]{Xueqi Cheng}
received the PhD degree in 2006, from the Institute of Computing Technology, Chinese Academy of Sciences (CAS), where he is currently a professor and the director of the CAS Key Laboratory of Network Data Science and Technology. His main research interests include network science, Web search and data mining, big data processing, distributed computing architecture, etc. He has published over 200 papers in reputable journals and conferences. He has won the Best Full Paper Runner-up Award in CIKM (2017), Best Student Paper Award in SIGIR (2012), and Best Paper Award in CIKM (2011). He currently serves on the editorial board for Journal of Computer Science and Technology, Journal of Computer, etc.
\end{IEEEbiography}

\vspace{-11mm}
\begin{IEEEbiography}
	[{\includegraphics[width=1in,height=1.25in,clip,keepaspectratio]{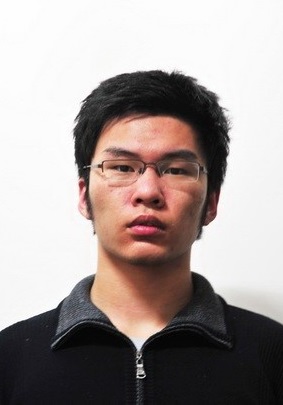}}]{Long Bai}
is a PhD student in the Institute of Computing Technology, Chinese Academy of Sciences. His current research interests include event knowledge graph, knowledge graph, etc. He has published papers in prestigious journals and conferences, including AAAI, EMNLP, etc.
\end{IEEEbiography}

\vspace{-15mm}
\begin{IEEEbiography}
	[{\includegraphics[width=1in,height=1.25in,clip,keepaspectratio]{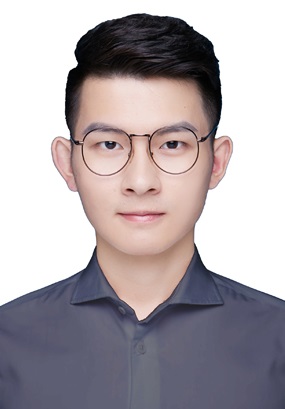}}]{Fujun Zhang}
is currently pursuing the master degree in the Institute of Computing Technology, Chinese Academy of Sciences. His current research interests include knowledge graph, information extraction, event extraction, etc.

\end{IEEEbiography}

\vspace{-15mm}
\begin{IEEEbiography}
	[{\includegraphics[width=1in,height=1.25in,clip,keepaspectratio]{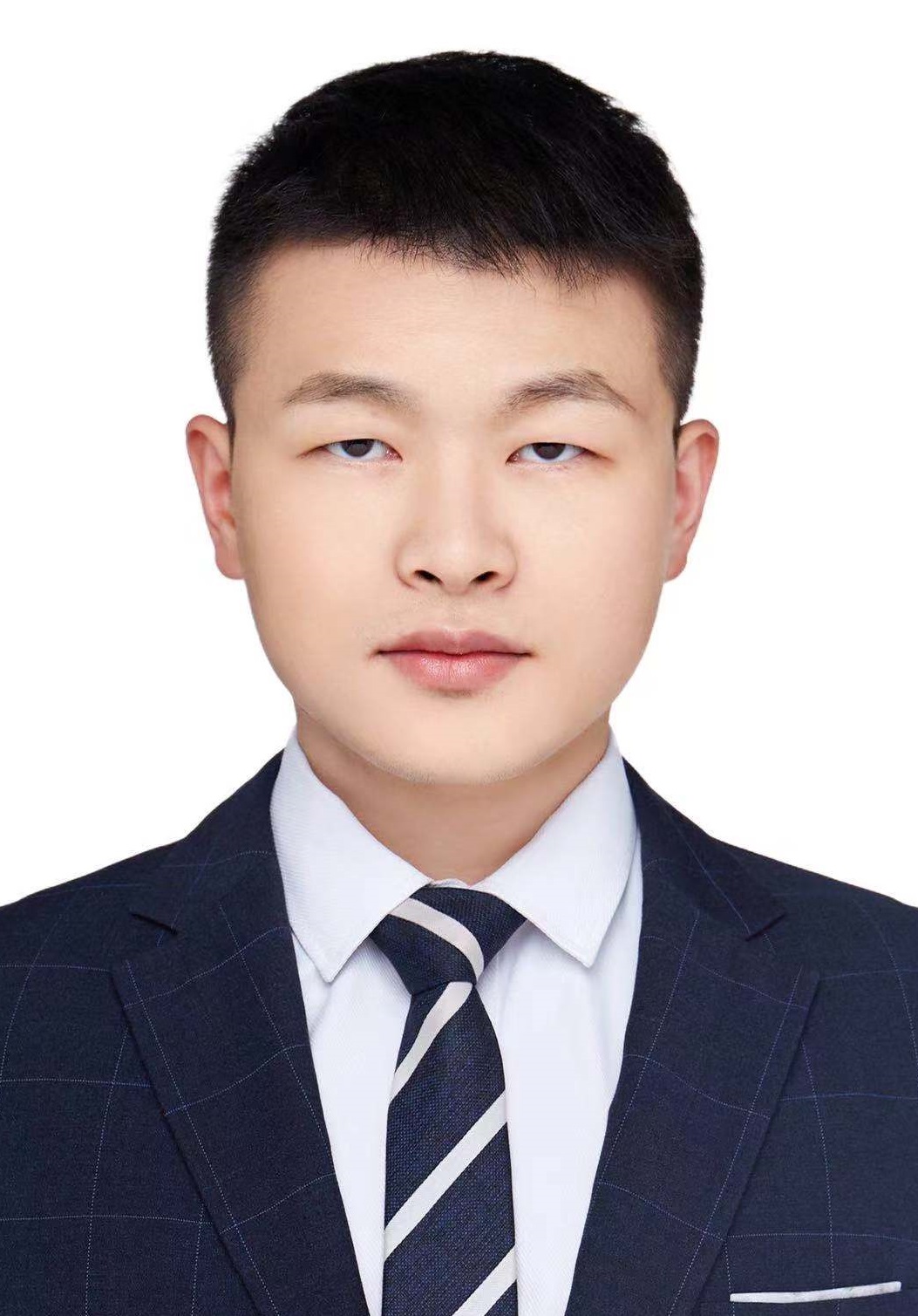}}]{Zixuan Li}
is a PhD student in the Institute of Computing Technology, Chinese Academy of Sciences. His current research interests include event knowledge graph, temporal reasoning, event reasoning, etc. He has published papers in prestigious conferences, including SIGIR, ACL, etc.
\end{IEEEbiography}

\vspace{-15mm}
\begin{IEEEbiography}
	[{\includegraphics[width=1in,height=1.25in,clip,keepaspectratio]{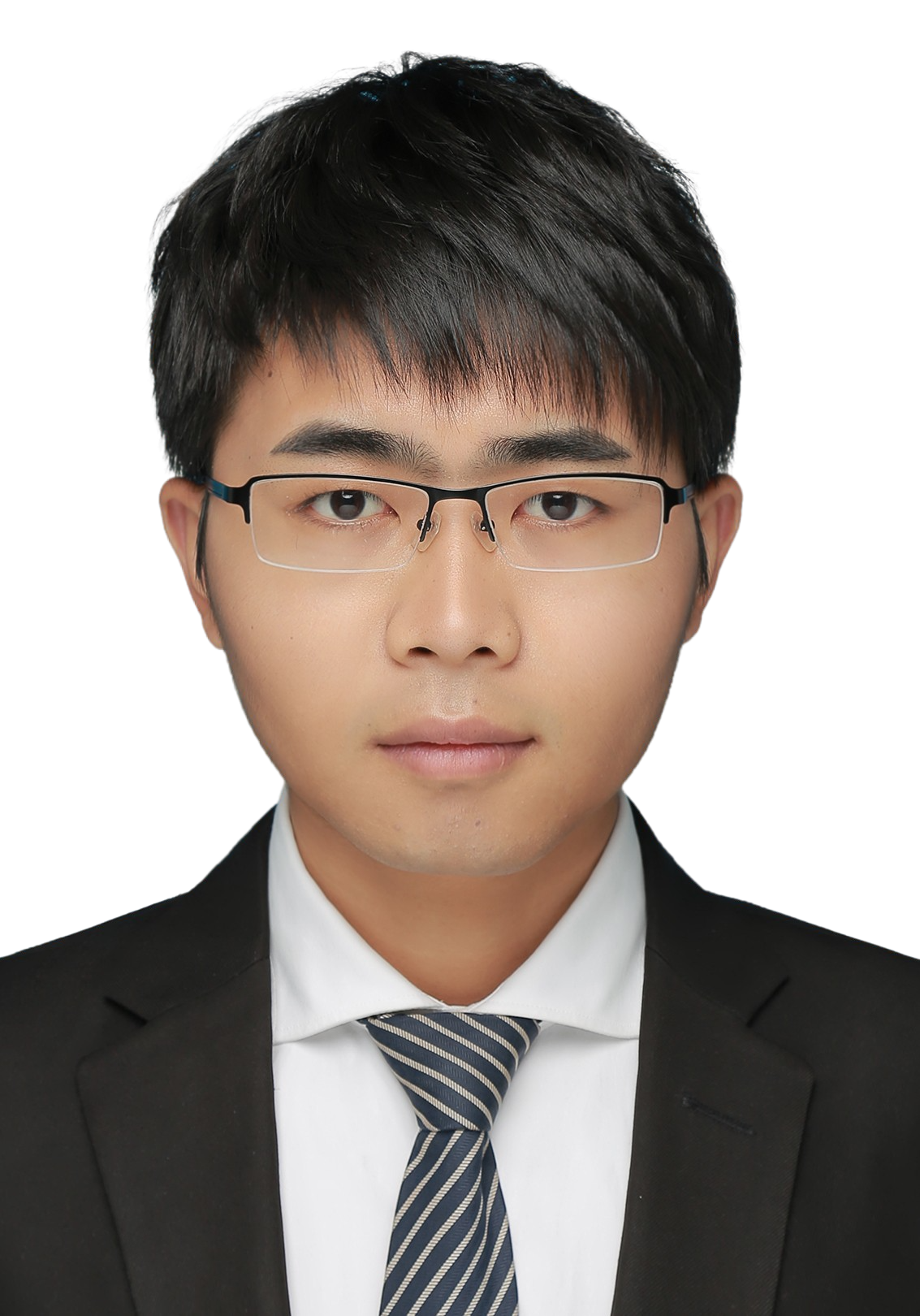}}]{Yutao Zeng}
received the master degree in Computer Applied Technology in 2018, from the Institute of Computing Technology, Chinese Academy of Sciences. He currently works at Tencent as a researcher of the Platform and Content Group. His research interests include event knowledge graph, event coreference resolution, etc.
\end{IEEEbiography}

\vspace{-14mm}
\begin{IEEEbiography}
	[{\includegraphics[width=1in,height=1.25in,clip,keepaspectratio]{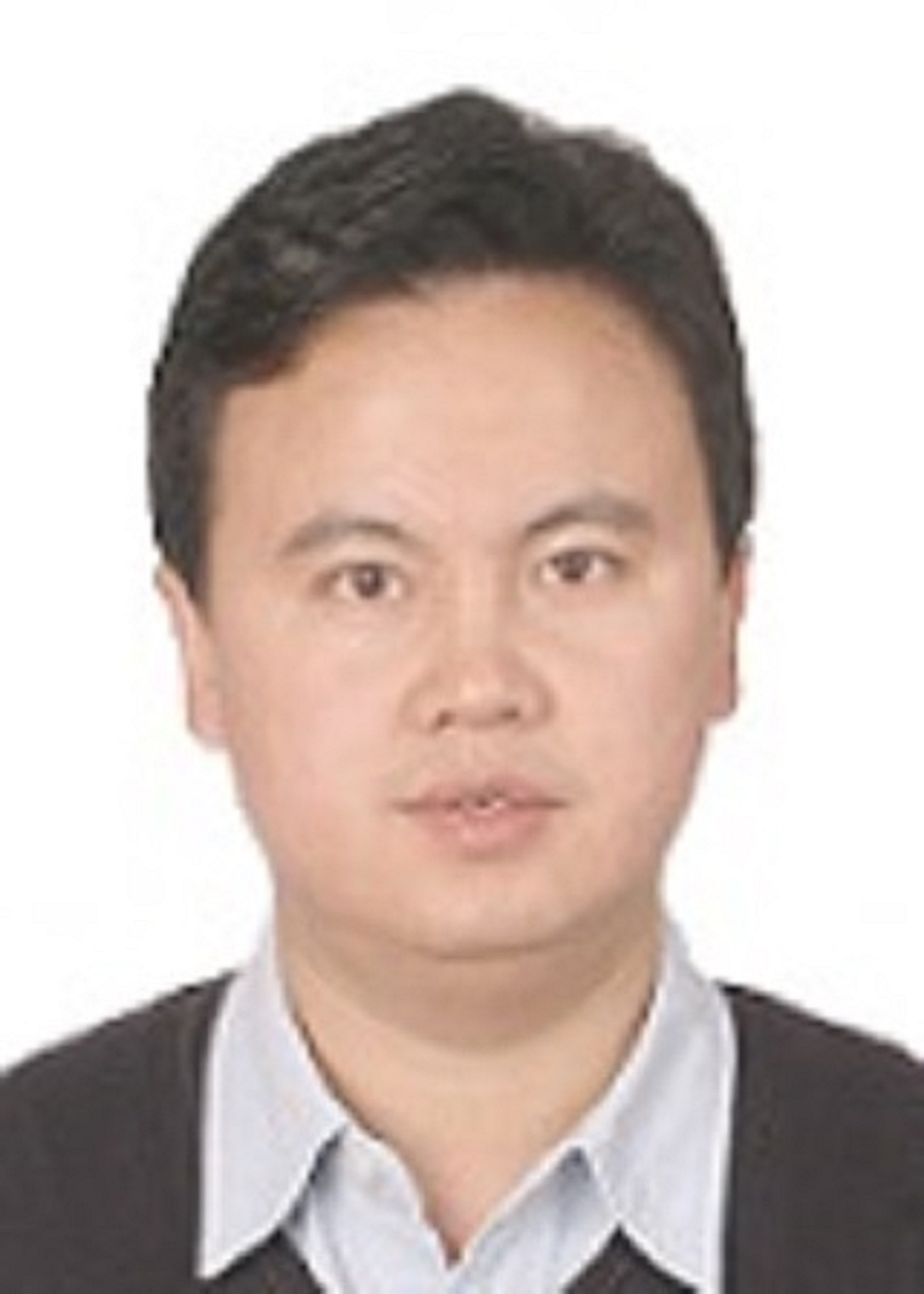}}]{Xiaolong Jin}
received the PhD degree in Computer Science from Hong Kong Baptist University in 2005. He is currently a professor in the Institute of Computing Technology, Chinese Academy of Sciences. His current research interests include knowledge graph, knowledge engineering, social computing, social networks, etc. He has published more than 200 papers in reputable journals and conferences. He has received the Best (Student/Academic) Paper Awards in ICBK (2017), CIT (2015), CCF Big Data (2015), AINA (2007), and ICAMT (2003).
\end{IEEEbiography}

\vspace{-11mm}
\begin{IEEEbiography}
	[{\includegraphics[width=1in,height=1.25in,clip,keepaspectratio]{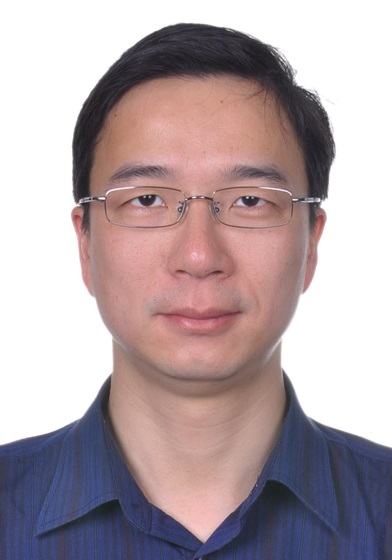}}]{Jiafeng Guo}
received the PhD degree in Computer Software and Theory in 2009, from the Institute of Computing Technology, Chinese Academy of Sciences (CAS), where he is currently a professor and the vice director of the CAS Key Laboratory of Network Data Science and Technology. He has worked on a number of topics related to Web search and data mining. His current research is focused on representation learning and neural models for information retrieval and filtering. He has won the Best Full Paper Runner-up Award in CIKM (2017), Best Student Paper Award in SIGIR (2012), and Best Paper Award in CIKM (2011).
\end{IEEEbiography}





\end{document}